\documentclass[runningheads]{llncs}

\usepackage{eccv}

\usepackage{eccvabbrv}

\usepackage{graphicx}
\usepackage{booktabs}
\usepackage{algorithm}
\usepackage{algpseudocode}
\usepackage{amsmath}

\usepackage[accsupp]{axessibility}

\usepackage{hyperref}

\usepackage{orcidlink}
\usepackage{caption}
\usepackage[table]{xcolor}
\usepackage{booktabs}
\usepackage{array}
\usepackage{amssymb}
\usepackage{multirow}
\usepackage[normalem]{ulem}
\useunder{\uline}{\ul}{}

\newcommand{\method}{$\mathbb{R}$-\texttt{GRPO}}
\begin{document}
 \newcommand{\sj}[1]{{\color{red}{[SJ:#1]}}}

\makeatletter
\newcommand\whline{\noalign{\ifnum0=`}\fi\hrule \@height 1.25pt \futurelet
	\reserved@a\@xhline}

\title{Beyond Time Shifts: Adapting Omni-LLM as \\a Reference-Free Evaluator for Generative Audio-Visual Models} 

\titlerunning{Adapting Omni-LLM as a Reference-Free AV Sync Evaluator}
\vspace{-4mm}
\author{Yijie Qian\inst{1,3}$^{\star\ddagger}$ \and
Juncheng Wang\inst{2}$^{\star}$ \and
Chao Xu\inst{3,4} \and
Huihan Wang\inst{2} \and
Yuxiang Feng\inst{1} \and
Yang Liu\inst{3,4} \and
Baigui Sun\inst{3,4} \and
Yong Liu\inst{1}$^{\dagger}$ \and
Shujun Wang\inst{2}$^{\dagger}$}

\authorrunning{Y.~Qian et al.}

\institute{Zhejiang University, Hangzhou, China \and
The Hong Kong Polytechnic University, Hong Kong, China \and
IROOTECH TECHNOLOGY, China \and
Wolf 1069 b Lab, Sany Group, China\\[0.6em]
\textnormal{$^{\star}$~Equal contribution.\quad
$^{\dagger}$~Corresponding authors.}\\[0.4em]
\textnormal{\small\{yijieqian@zju.edu.cn, wjc2830@gmail.com\}$^{\star}$,\quad
\{yongliu@iipc.zju.edu.cn,\\ shu-jun.wang@polyu.edu.hk\}$^{\dagger}$}}

\maketitle
\begingroup\renewcommand\thefootnote{}\footnotetext{$^{\ddagger}$~This work was conducted in collaboration with IROOTECH TECHNOLOGY.}\addtocounter{footnote}{-1}\endgroup

\vspace{-6mm}
\begin{abstract}
  As audio-visual generative models evolve into world simulators, cross-modal synchronization stands as a critical proxy for assessing the consistency of world dynamics and causality in generated content. However, existing evaluation metrics presume structural correctness, reducing synchronization to mere temporal alignment. Consequently, they fail on generative outputs, especially when exhibiting structural hallucinations and asymmetric cross-modal relations, which currently \textbf{mandate expert human annotation to assess synchronization.} This dependency introduces a critical paradox: \emph{human evaluators rely on relative, reference-dependent comparisons, whereas automated metrics require reference-free, absolute scalars.} We resolve this paradox by proposing a framework that distills relative human perception into a continuous, globally consistent metric. First, we introduce SynthSync, a dataset of generative failures ranked via pairwise human annotations. Second, we adapt the Omni-LLM equipped with a continuous latent projection to translate relative human rankings into continuous absolute values. Third, we propose Real-Valued Group Relative Policy Optimization ($\mathbb{R}$-GRPO) to internalize the global causal structure of synchronization via listwise score distributions. Empirically, our metric achieves state-of-the-art human preference alignment. We leverage this estimator to establish a standardized benchmark, advancing AV-Gen assessment from low-level signal correlation to visually grounded causality. Project page: \href{https://chenhaoqcdyq.github.io/BeyondTimeShifts}{https://chenhaoqcdyq.github.io/BeyondTimeShifts}
  \keywords{Joint Audio-Video Generation \and Audio-Visual Synchronization \and Omni-LLM}
\end{abstract}

\vspace{-8mm}
\section{Introduction}
\label{sec:intro}

\begin{figure}[t]
    \centering
    \includegraphics[width=\linewidth]{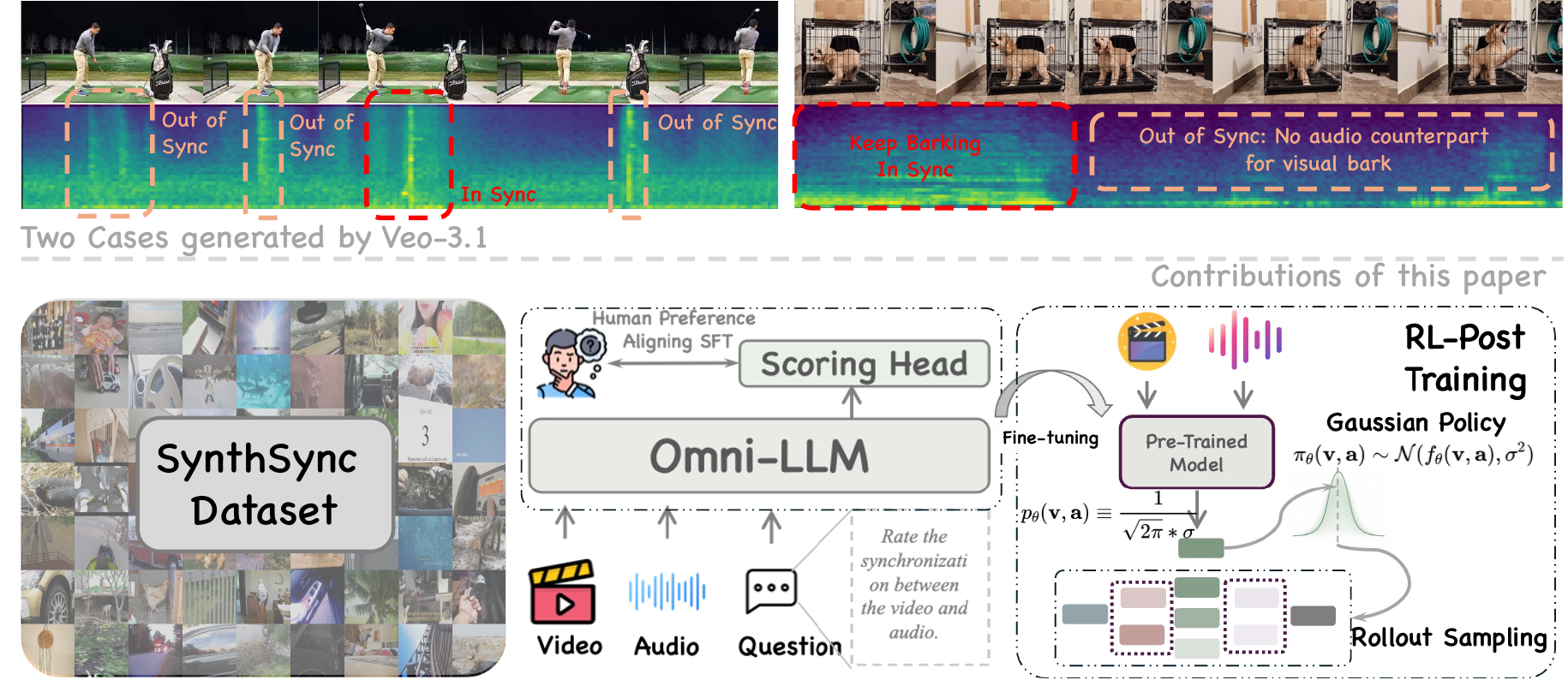}
    \caption{\textbf{Upper row}: Two Veo-3.1~\cite{veo3}-generated audio-video samples with evident mismatches between visual events and audio, showcasing structural hallucinations where the synthesized impact sounds (e.g., the golf strike) exhibit semantic and physical textures inconsistent with the visual cues, and a breakdown in asymmetric relationships where the visual (e.g., dog barking) lacks corresponding audio signal. \textbf{Lower row}: Our synchronization evaluation pipeline. We first introduce \emph{SynthSync} (left), a dataset of authentic generative synchronization failures with pairwise preference annotations. We then fine-tune an omni-LLM as a reference-free continuous synchronization scorer using a new training paradigm (middle), followed by reinforcement post-training to better capture the global causal structure of synchronization judgments (right).}
    \label{fig:1}
\end{figure}

Driven by systems like Sora-2~\cite{openai2025sora2,sora1}, Veo-3~\cite{veo2,veo3}, and Seedance-2~\cite{SeD15P,seedance2}, audio-visual generative models~\cite{MOVA,Kling3,UniVerse,JavisDiT,OmniCustom} are evolving from visual synthesizers into nascent ``world simulators''~\cite{sora1}.
While these models achieve single-modality fidelity, the synchronization of cross-modal events remains fragile (shown in upper row of Fig.~\ref{fig:1}). Such failures matter not only for perceptual realism, but also as a key observable proxy for whether a model has learned how actions and sounds relate in the real world.

However, measuring this new capability with current evaluation paradigms reveals a fundamental mismatch. Traditional metrics, such as onset-based synchronization accuracy~\cite{AVAlign}, cross-modal contrastive similarity~\cite{ImageBind}, and offset-detection scores~\cite{iashin2024synchformer,Alignnet}, are all grounded in a structural correctness assumption: they presume that audio and video content are drawn from reality, semantically matched, and require only simple temporal alignment. Nevertheless, modern generative outputs~\cite{openai2025sora2,veo3,seedance2} often break this assumption entirely. Instead of clean temporal shifts~\cite{iashin2024synchformer}, their failures frequently involve structural hallucinations (e.g., generating a sharp glass-shattering sound when a plastic cup falls onto a soft carpet), asymmetric relationships~\cite{JavisDiT} (e.g., a visual action implying a sound that never occurs), and temporally diffuse events that defy simple onset detection~\cite{AVAlign} (e.g., the continuous pouring of liquid where the acoustic pitch fails to dynamically rise with the visual fluid level). 

Because standard algorithmic matching cannot resolve these unpredictable structural errors, establishing a reliable ground truth for audio-visual synchronization fundamentally requires \textbf{human cognition}. The high perceptual fidelity of modern generative artifacts limits ordinary rule-based discrimination, which often fails to distinguish subtle structural violations from overall audio-visual quality. Consequently, expert human preference serves as the necessary oracle, shifting the evaluation paradigm from signal-level temporal offsets to genuine causal perception.

Translating this human intuition into a scalable automated metric, however, introduces a central methodological \textbf{paradox}. When evaluating complex generative errors, human annotators naturally rely on relative comparisons—judging which of two flawed audio tracks fits a video better is easier than directly scoring a single one in isolation. In contrast, a practical and deployable evaluation metric must be reference-free, assessing a single generated video-audio pair in the wild and outputting an absolute, continuous score. This disconnect between relative human perception and the need for an absolute measurement motivates our core research question: \textbf{How can we distill the complex, comparative nature of human judgment regarding synchronization into a continuous, reference-free, and globally consistent automated metric?}

In this paper, we systematically resolve this paradox through a three-stage methodological progression, as shown in the lower row of Fig.~\ref{fig:2}. First, to ground the subjective nature of human judgment, we construct the SynthSync dataset. Instead of using artificial time shifts~\cite{iashin2024synchformer}, we curate authentic generative failures by sampling outputs from 10 state-of-the-art video-to-audio models (see Section~\ref{sec:exp} for complete list) given \textbf{the same visual anchor}. This allows us to 1) simulate the complex topology of synchronization errors in modern AV-Gen systems; and 2) collect fine-grained pairwise rankings ($<\mathbf{v},\mathbf{a}_1>$ \emph{vs.} $<\mathbf{v},\mathbf{a}_2>$) over expert human annotations. 

 Second, to enable reference-free scoring, we architecturally redesign an Omni-LLM. Specifically, we inherit the structural correctness knowledge embedded in Omni-LLM and adapt it for scoring tasks by replacing its discrete language head with a continuous projection head from the latent manifold. We initialize this scorer using a curriculum-based preference learning strategy, mapping relative human rankings into absolute scalar values via a Bradley-Terry-Luce formulation~\cite{tutz1986bradley}. Finally, to enforce global consistency beyond local pairwise comparisons, we introduce a novel reinforcement learning framework: \textbf{$\mathbb{R}$-GRPO (Real-Valued Group Relative Policy Optimization)}. By treating the estimator's output as a stochastic Gaussian policy, we optimize the entire listwise distribution of scores against the ground truth topology, allowing the model to internalize the global causal structure of synchronization.

Empirically, extensive experiments demonstrate that our continuous estimator achieves state-of-the-art (SOTA) alignment with human preference in causal evaluation. Leveraging this validated metric, we establish a comprehensive benchmark, SyncBench, for contemporary AV-Gen models, providing the field with a standardized tool to assess the physical grounding of nascent world simulators.
 
Our contributions are fourfold:
\begin{itemize}
    \item \textbf{SynthSync:} The first dataset of authentic generative synchronization failures, moving the field beyond artificial time-shift benchmarks.
    \item \textbf{Continuous Omni-LLM Evaluator:} A novel architecture that repurposes multimodal foundation models for continuous, reference-free causal consistency scoring.
    \item \textbf{$\mathbb{R}$-GRPO Framework:} A scalable reinforcement learning algorithm that aligns continuous latent policies with discrete human preferences, establishing a new standard for physically grounded generative evaluation.
    \item \textbf{SOTA Alignment and SyncBench:} We demonstrate SOTA performance in human preference alignment and establish a standardized benchmark for evaluating AV-Gen models.
\end{itemize}

\section{Related Work}
\label{sec:literature}

\subsection{Audio-Video Joint Generation}
Early audio-visual synthesis commonly adopted a cascaded design, in which video generation~\cite{cui2026lol,km2026phyeduvideo,brokman2026training,wan2025wan,zhang2026reward,zhu2026causal,huang2025self,cuiself,qian2025thinkmovelatentmotion,ma2026fastvmteliminatingredundancy,feng2026newtonagenticplanningphysically,xu2023high,xu2024facechain} and audio generation~\cite{haji2026taming,wang2025language,ETTA,SAO,Fugatto,wang2026guided} were treated as separate stages. A silent video was first generated, and a downstream video-to-audio~\cite{AudioX,MMAudio,MelQCD,FoleyCrafter,Cafa,Selva,Thinksound,PrismAudio,HunYuanFoley,Lova,V2ALDM} (V2A) model was then used to synthesize the soundtrack.

More recent work moves toward native joint generation. A prominent design is the Dual-Branch Diffusion Transformer, adopted by the Seedance family~\cite{SeD15P,seedance2} and related systems~\cite{JavisDiT,OmniCustom,javisdit++,ALIVE,LTX2,MOVA,Klear,Kling3, UniVerse,ma2026easyvfxfrequencydrivendecoupling,Ma_2026_CVPR,weng2025audiosyncvideogenerationmultistream,zheng2026aligning,song2025syncphony}, where audio and video are modeled in parallel by modality-specific transformer branches coupled through cross-modal interaction modules. By contrast, closed-source systems such as Sora-2 and the Veo-3 series provide little public technical detail, leaving their underlying solutions unavailable for open analysis.

\subsection{Audio-Video Synchronization Evaluation}
Audio-visual synchronization has traditionally been measured by algorithmic metrics such as SyncNet~\cite{Syncnet}, PEAVS~\cite{goncalves2024perceptual}, and AlignNet~\cite{Alignnet}. SyncNet evaluates alignment mainly through global temporal offsets between speech audio and lip motion. PEAVS improves perceptual modeling, but generative evaluation in practice is still often dominated by unimodal fidelity measures such as FAD and FID, which do not explicitly model cross-modal temporal dependence. AlignNet, meanwhile, is designed for human motion--audio alignment and is not readily transferable to general open-domain scenes.

These approaches become increasingly inadequate for modern generative outputs. They typically assume that audio and video are structurally correct up to a simple temporal shift, that synchronization is expressed by sparse transient events such as lip closures or percussive onsets, and that cross-modal similarity can be captured by symmetric matching. Such assumptions break down in realistic generative failures, where errors are often semantic, asymmetric, and temporally diffuse.

Recent methods increasingly rely on pretrained cross-modal encoders, e.g., ImageBind~\cite{ImageBind}, to assess synchronization~\cite{JavisDiT}. However, such models are still constrained by pretraining objectives that emphasize global and compressed cross-modal correspondence rather than fine-grained causal timing. Related methods such as AV-Align further formulate synchronization as an onset- or event-matching problem, which is less suitable for continuous phenomena such as flowing water, fire, or explosions, where synchronization is gradual rather than event-localized.

\subsection{LLM-as-Judge}

\paragraph{Evaluating Language Responses.}
LLM-as-Judge effectiveness hinges on evaluator design rather than model scale. Prior work mitigates bias via structured prompting\cite{mohammadkhani2025checklist}, dynamic multi-agent judges~\cite{cao2025multi,zhang2025agent}, and contrastive scoring~\cite{wang2025contrastscore}; or enhances reasoning through pairwise comparison~\cite{shibata2025lces}, continuous ranking~\cite{mu2025evaluate}, and RL-based training~\cite{whitehouse2025j1}. However, these overlook temporally grounded cross-modal generation, where errors are asymmetric, structurally unconstrained, and ill-suited to text-centric evaluation criteria.

\paragraph{Evaluating Multi-Modal Synthesis.}

LMM/VLM-based evaluators primarily target image/video quality via discrete text-defined scoring\cite{wu2023q}, pairwise-to-score conversion~\cite{zhu2024adaptive,zhao2025reasoning}, and RL-based quality understanding~\cite{QInsight,wu2025visualquality,zhang2025vq}, with extensions to document images~\cite{gao2025deqa}, and diagnostic flaw maps for text-to-image generation~\cite{guo2025imagedoctor}. Yet these address perceptual quality or aesthetics, not the directed causal-semantic consistency central to audio-visual synchronization—where visual events must yield physically coherent acoustic outcomes. This gap motivates our specialized dataset, supervision signal, and continuous RL-based judge formulation tailored to authentic V2A synchronization failures.

\section{Method}
\label{sec:method}

\begin{figure}[t]
    \centering
    \includegraphics[width=\linewidth]{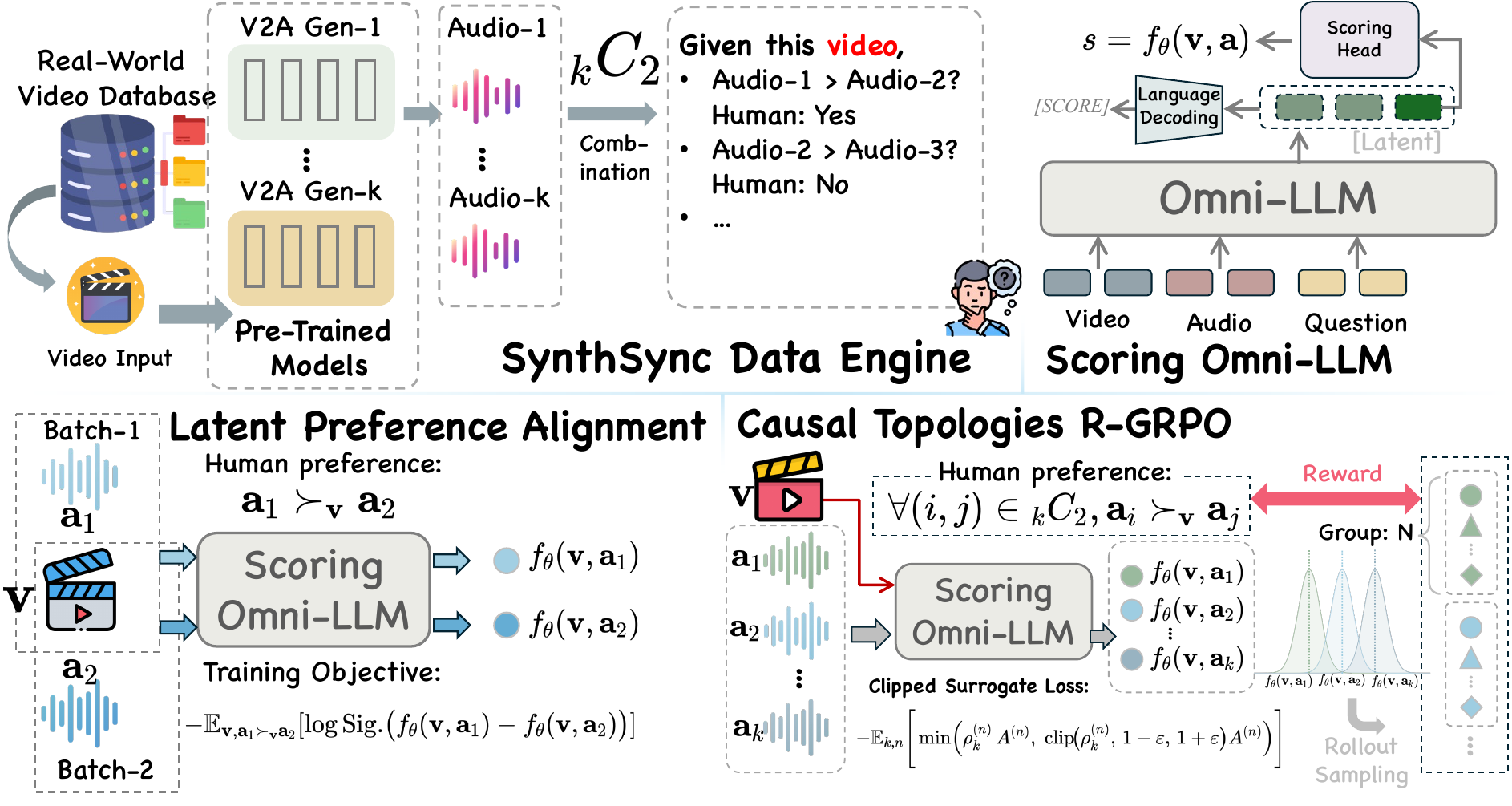}
    \caption{\textbf{Overview of our framework for audio-video synchronization evaluation.} \underline{Upper left:} We construct \emph{SynthSync} by applying $k$ video-to-audio (V2A) generators to each real-world video and collecting human pairwise annotations of which generated audio is better synchronized with the video. \underline{Upper right:} The architecture of our scoring omni-LLM, adapted from an omni-LLM backbone. \underline{Lower left:} The scorer is first trained with a Bradley--Terry--Luce objective on pairwise preferences. \underline{Lower right:} We then apply \method~for reinforcement post-training, improving preservation of the causal structure underlying synchronization evaluation.}
    \label{fig:2}
\end{figure}

Evaluating audio-visual synchronization in generative models demands perceiving complex, structurally unconstrained semantic errors. 
Operationalizing this requires resolving a central methodological paradox. \underline{On one hand}, human annotation of generative errors is inherently relative---annotators \emph{require an anchor} to make reliable psychophysical judgments. 
\underline{On the other hand}, a practical evaluation metric must be \emph{absolute and reference-free}---capable of observing a single generated video-audio pair in the wild and outputting a continuous scalar. 
Our methodology systematically dismantles this tension through a novel dataset curation strategy (SynthSync), a continuous Omni-LLM architecture, and a two-stage topological alignment paradigm, with Fig.~\ref{fig:2} depicting the framework.

\subsection{The Annotation Paradox and the SynthSync Dataset}
Developing a robust evaluator requires a human-aligned supervisory signal. However, the community currently lacks a dataset capable of benchmarking the true nature of modern generative failures. Existing evaluation protocols rely almost exclusively on artificial linear offsets—mechanically shifting ground-truth audio forward or backward. This approach fundamentally misrepresents the generative landscape, heavily rewarding models that merely detect offset artifacts rather than those that understand physical causality.

To bridge this gap, we curate \textbf{SynthSync}, the first dataset dedicated to authentic generative synchronization failures. Given an underlying real-world visual stream $\mathbf{v}$, we generate multiple distinct acoustic tracks by passing the video through $K$ distinct, cutting-edge video-to-audio (V2A) foundation models, forming a candidate pool $\mathcal{A}(\mathbf{v})=\{\mathbf{a}_1,...,\mathbf{a}_K\}$. This strategy offers two critical advantages over traditional dataset construction:
\begin{enumerate}
    \item \textbf{Simulating Authentic Structural Chaos}: Unlike manual temporal shifting, multi-model generation natively surfaces the complex, non-linear failure modes endemic to modern AV-Gen systems. This includes semantic hallucinations (e.g., a dog barking with a bird’s chirp), temporal diffusion (continuous water flow without distinct onsets), and severe causal omissions.
    \item \textbf{Visual Anchoring for Psychophysical Consistency}: Given a single heavily hallucinated audio-video pair, attempting to annotate an absolute synchronization ``score'' without references is an ill-posed psychophysical task, introducing extreme inter-annotator variance. By setting the underlying video $\mathbf{v}$ as a fixed causal anchor, annotators are instead tasked with a comparative judgment: \emph{does candidate $\mathbf{a}_i$ better match the video than candidate $\mathbf{a}_j$?} This grounds the subjective nature of multimodal physics into strictly mathematically operable partial orderings ($\mathbf{a}_i \succ_\mathbf{v} \mathbf{a}_j$).
\end{enumerate}

\subsection{The Inference Requirement: Reference-Free Continuous Scoring}
\label{sec:continuous_evaluators}
While SynthSync relies on relative rankings to guarantee human consistency, a viable Omni-LLM evaluator must be structurally \textbf{reference-free} at inference---capable of emitting a standalone metric for any $(\mathbf{v}, \mathbf{a})$ pair. Furthermore, the dominant ``LLM-as-a-Judge'' paradigm typically prompts models to emit discrete linguistic categories (e.g., ``Good'', ``Poor''). In the context of audio-visual physics, fine-grained temporal and semantic variations defy clean mapping into broad linguistic buckets; \emph{doing so inevitably introduces severe quantization artifacts and lexical biases}.

Rather than forcibly categorizing human perception, we bypass token decoding altogether. Given a video $\mathbf{v}$ and a generative candidate audio $\mathbf{a}$, we extract a scalar quality representation directly from the \textbf{underlying causal manifold of the model}. We force it to output a persistent inference token \texttt{[SCORE]} into the multimodal context sequence. Letting $\mathbf{h}_{\text{score}} \in \mathbb{R}^d$ denote the model's ultimate hidden representation corresponding to this token, we append a specialized continuous projection head $\text{MLP}_\phi$:
\begin{equation}
    s = f_\theta (\mathbf{v}, \mathbf{a}) = \text{MLP}_\phi \left(\mathbf{h}_{\text{score}}\right) \in \mathbb{R}
\end{equation}

This parameterization cleanly strips away language-generation stochasticity, transmuting the Omni-LLM into a fully differentiable mapping $f_\theta$ optimized exclusively for relative continuous scaling.

\subsection{Stage I: Latent Preference Alignment via Curriculum}
\label{sec:stage_i}
We are now faced with an optimization gap: our model $f_\theta$ must output an absolute, reference-free scalar $s$, but our ground truth consists exclusively of relative preference boundaries. Standard Mean Squared Error (MSE) regression introduces active harm by compressing highly contextual psychophysical phenomena into rigid numeric bounds. 

We consequently abandon regression-based absolute distance, directly optimizing the global topological rank mapping via the Bradley-Terry-Luce (BTL)~\cite{tutz1986bradley} formulation. For each paired ranking $\mathbf{a}_i \succ_\mathbf{v} \mathbf{a}_j$, we minimize the localized loss over the logit decision boundary:
\begin{equation}
    \label{eq:BTL}
    \mathcal{L}_{\text{BTL}}(\theta) = -\mathbb{E}_{\mathbf{v}, \mathbf{a}_i \succ_\mathbf{v} \mathbf{a}_j} \left[ \log \text{Sig.} \big(f_\theta(\mathbf{v}, \mathbf{a}_i) - f_\theta(\mathbf{v}, \mathbf{a}_j)\big) \right],
\end{equation}
where Sig. represents the sigmoid activation. Although supervised only \emph{within} each anchor, fitting all margins through a single shared $f_\theta$ induces a globally consistent implicit value space---rather than a per-anchor calibration---making the absolute scalar $s=f_\theta(\mathbf{v},\mathbf{a})$ directly usable as a reference-free metric across anchors at inference, as formalized in the reward-modeling analysis of DPO~\cite{rafailov2023direct}.

\textbf{Curriculum Pacing:} Uniformly sampling pairwise preferences risks gradient collapse due to the massive variance in generative errors---ranging from obvious semantic hallucinations to nuanced sub-frame asynchronies. To stabilize optimization, we structure the learning trajectory using a dynamically annealed curriculum. Given a set of pairwise annotations for a single video, we first establish a global ordinal ranking of the candidate audios by aggregating their empirical win-rates (the total number of head-to-head victories per candidate). Let $\delta(i, j)$ denote the distance between two candidates in this derived global ranking. We initialize training by exclusively sampling pairs with large rank discrepancies---for instance, contrasting the top-ranked candidate (No.~1) with a severely degraded one (No.~9)---ensuring a strong, unambiguous initial gradient signal. As training progresses, we decay a threshold constraint $\delta(i, j) \ge \gamma(t)$, safely uncoiling the margin to expose the model to finer, microscopic comparisons (e.g., No.~5 vs.~No.~6) and steadily refining its perceptual boundaries.

\subsection{Stage II: Causal Topologies via $\mathbb{R}$-GRPO}
\label{sec:stage_ii}

\textbf{From Local Pairs to Global Lists:} Pairwise learning of Eq.~\ref{eq:BTL} suffers from \textbf{metric myopia}: establishing local margins ($\mathbf{a}_i \succ_\mathbf{v} \mathbf{a}_j$) does not guarantee a globally transitive scale. Consequently, the model may satisfy individual constraints while producing distorted, non-linear score distributions that fail on holistic ranking metrics. To enforce global topological consistency, we transcend pairwise proxies $\{i,j|\forall i,j\le K,i\ne j\}$ and optimize the entire listwise distribution $\{1,...,K\}$ directly using reinforcement learning (RL).

\noindent\textbf{From Deterministic Regression to Stochastic Policy.} However, applying standard RL presents a structural paradox: policy gradient methods fundamentally require \textbf{stochastic exploration} over a probability distribution to estimate advantages, whereas our evaluator $f_\theta$ is a \textbf{deterministic regressor} outputting a fixed scalar. To bridge this gap, we propose $\mathbb{R}$-\textbf{GRPO} (Real-Valued Group Relative Policy Optimization). 
We reparameterize the deterministic scalar output not as a static value, but as the central manifold $\mu_\theta$ of a Gaussian policy, injecting the necessary uncertainty to enable differentiable policy optimization.

\noindent\textbf{Exploratory Listwise Rollouts:} Rather than treating the final projection $f_\theta(\mathbf{v}, \mathbf{a}_k)$ as an inflexible terminal state, we reconstruct it as the central manifold $\mu_\theta$ of an unconstrained exploratory Gaussian policy space. During optimization rollouts for a single anchor video $\mathbf{v}$, the current deterministic estimator issues base continuous values $\{\hat{s}_1, \dots, \hat{s}_K\}$. To trigger dynamic evaluation environments, we sample $N$ parallel execution paths by systematically perturbing the scores within a designated active-exploration variance $\sigma^2$:
\begin{equation}
    \hat{s}_{k}^{(n)} = \hat{s}_{k} + \epsilon_{k}^{(n)}, \qquad \epsilon_{k}^{(n)} \sim \mathcal{N}(0, \sigma^2)
\end{equation}
This formulation projects the configuration group $\mathcal{G}_i = \{\hat{\mathbf{s}}^{(1)}, \dots, \hat{\mathbf{s}}^{(N)}\} \subset \mathbb{R}^K$ directly into the continuous domain, natively unlocking deep state-space trajectories that bypass auto-regressive decoder constraints.

\noindent\textbf{Composite Causal-Semantic Rewards:} To guarantee the preservation of global causal topologies across the candidate space, we sample all of generated \emph{audios} from the same \emph{video}, and obtain their scores, which is used to against the ground-truth reference ranking of corresponding audio using a holistic Composite Ranking Reward $r \in [0, 1]$. This formulation directly optimizes the structural consistency of the entire distribution list, which evaluates the sequence natively through three complementary topological lenses: 1) \textit{structural attenuation}, 2) \textit{monotonic topological order}, and 3) \textit{extreme target bounding}. (See Appendix for detailed formulation.)

\noindent\textbf{Group Bounded Policy Objectives:} We adapt the GRPO framework~\cite{guo2025deepseek} to continuous latent spaces, leveraging group-relative normalization to eliminate the need for a separate value critic. For a sampled ensemble of $N$ trajectories, we compute variance-reduced advantages $A^{(n)}$ relative to the local group mean $\bar{r}$:
\begin{equation}
    A^{(n)} = \frac{r^{(n)} - \bar{r}}{\text{std}\big(\{r^{(n)}\}_{1}^{N}\big) + \epsilon}
\end{equation}
To ensure monotonic policy improvement, we constrain gradient updates using a proximal trust region. Since our policy is parameterized as a Gaussian $\pi_\theta(\cdot|\mathbf{v},\mathbf{a}) = \mathcal{N}(\mu_\theta, \sigma^2)$, the importance sampling ratio $\rho_{k}^{(n)}$ admits a closed-form analytical solution:
\begin{equation}
    \rho_{k}^{(n)} = \frac{\pi_\theta \left(\hat{s}_{k}^{(n)}\right)}{\pi_{\theta_{\text{old}}}\left(\hat{s}_{k}^{(n)}\right)} = \exp \!\left(-\frac{\left(\hat{s}_{k} - \hat{s}_{k}^{(n)}\right)^2}{2\sigma^2}\right) 
\end{equation}

The final objective $\mathcal{L}_{\mathbb{R}\text{-GRPO}}$ integrates the clipped surrogate loss with a standard KL divergence penalty to strictly regularize the updated policy against the reference semantic backbone $\pi_{\theta_{\text{ref}}}$:
\begin{align}
    \mathcal{L}_{\mathbb{R}\mathtt{-GRPO}} = &-\mathbb{E}_{k,n}\Bigg[ \min \!\Big(\rho_{k}^{(n)}\, A^{(n)},\; \text{clip}\!\big(\rho_{k}^{(n)},\, 1-\varepsilon,\, 1+\varepsilon \big) A^{(n)}\Big) \Bigg] \\ &+ \lambda_{\text{KL}}\; \mathbb{E}_{k}\!\Bigg[\frac{\sigma^2 + \left(\hat{s}_{k} - \hat{s}_{k}^{\,\text{ref}}\right)^2}{2\sigma^2} - \frac{1}{2}\Bigg]
\end{align}

This formulation allows the Omni-LLM to optimize macroscopic structural constraints directly in continuous space, remaining robust to the arbitrary modality perturbations inherent in generated streams.

\section{Experiments}
\label{sec:exp}
\subsection{SynthSync Dataset}
\label{subsec:dataset}

\noindent\emph{Dataset Curation.} We construct the dataset by generating acoustic tracks for real-world visual streams using 10 state-of-the-art video-to-audio (V2A) foundation models. To mitigate data contamination, source videos are drawn exclusively from the test splits of VGGSound\cite{chen2020vggsoundlargescaleaudiovisualdataset} and FoleyBench\cite{dixit2025foleybenchbenchmarkvideotoaudiomodels}. Rigorous human 

\noindent
\begin{minipage}[t]{0.65\linewidth}
  \vspace{0pt} 
  filtering excludes clips with occluded sound sources, confounding background audio, or disruptive editing that violates physical causality. This pipeline isolates complex failure modes endemic to modern AV-Gen systems, including semantic hallucinations, temporal diffusion, and causal omissions.
  
\end{minipage}
\hfill
\begin{minipage}[t]{0.32\linewidth}
  \vspace{0pt}
\vspace{-5mm}
\captionof{table}{Detailed list for V2A models.}
  \renewcommand{\arraystretch}{1.15}{
    \resizebox{\textwidth}{!}{
    \begin{tabular}{c|c}
    \whline
    AudioX~\cite{AudioX}       & FoleyCrafter~\cite{FoleyCrafter} \\ \hline
    CAFA~\cite{Cafa}         & FoleyControl~\cite{rowles2025foley} \\ \hline
    HunYuanFoley~\cite{HunYuanFoley} & LoVA~\cite{Lova}         \\ \hline
    MelQCD~\cite{MelQCD}       & MMAudio~\cite{MMAudio}      \\ \hline
    Selva~\cite{Selva}        & VTA-LDM~\cite{V2ALDM}      \\ \whline
\end{tabular}
}}
\end{minipage}

\noindent\emph{Annotation Protocol.} To establish a reliable ground truth for synchronization, we bypass absolute scoring and instead rely on relative comparative judgments. We recruited an expert annotation pool of 20 individuals, dividing them into three independent groups. Each group provided complete pairwise annotations for the entire dataset, judging which candidate audio track ($a_{i}$ or $a_{j}$) better matched the fixed visual anchor. Following the collection of these pairwise preferences, we calculated the empirical win-rates (total head-to-head victories) for each audio candidate. By aggregating these win-rates across all three annotation groups ($2,267\times {}_{10}C_{2}\times3\approx 306K$ annotations), we derived a robust, final global ordinal ranking for all generated audios associated with the same video. This strictly operable partial ordering forms the ground-truth topology for our continuous latent alignment.

\subsection{Evaluation Setup}
\label{subsec:evaluation_setup}

\noindent\emph{Benchmarks.} We train and evaluate our continuous estimator exclusively on the proposed SynthSync dataset, assessing audio-visual temporal synchronization using evaluation metrics such as NDCG\cite{jarvelin2002cumulated}, Kendall's $\tau$\cite{kendall1938new}, and Pairwise Accuracy (Pair-ACC)\cite{ouyang2022training} to measure both structural causal coherence and fine-grained alignment. Detailed descriptions of the dataset and curation pipeline are provided in Section~\ref{subsec:dataset}.

\noindent\emph{Implementation Details.} Built on Qwen2.5-Omni-3B, our model replaces the discrete language head with a continuous projection MLP initialized via pairwise SFT. Inputs are standardized to 5s duration and $140\times140$ resolution at 12 FPS. We employ \method~with listwise batches ($K=6$), sampling 12 Gaussian variants per output ($\sigma^2=2.5$). Optimization uses AdamW (lr=$1e-6$) with PPO clipping ($\epsilon=0.2$) and KL penalty ($\beta=0.001$) over 100 epochs. See Appendix for further hardware and training specifications.

\subsection{Main Results}
\begin{table}[t]
\centering
\caption{Quantitative comparison of our proposed metric against baseline evaluation methods on the SynthSync benchmark. The baselines include zero-shot off-the-shelf Omni-LLMs, non-LLM synchronization metrics, and explicitly trained Omni-LLM evaluators. Best results are in bold, and second-best are underlined.}
\renewcommand{\arraystretch}{1.15}{
    \resizebox{\textwidth}{!}{
\begin{tabular}{cccccccc}
\whline
\multicolumn{1}{c|}{\multirow{2}{*}{\textbf{Method}}} & \multicolumn{2}{c|}{\textbf{Cumulative Gain} }                  & \multicolumn{2}{c|}{\textbf{Correlation}}                       & \multicolumn{2}{c|}{\textbf{Confusion}    }                   & \textbf{Pair-wise}      \\ \cline{2-8} 
\multicolumn{1}{c|}{}                        & NDCG            & \multicolumn{1}{c|}{MRR}             & Kendall         & \multicolumn{1}{c|}{Spearman}        & Top-1 Acc.     & \multicolumn{1}{c|}{Bottom-1 Acc.}  & Pair Acc.      \\ \whline
\multicolumn{8}{c}{\texttt{Off-the-shelf Omni-LLM}}                                                                                                                                                                                             \\ \hline
\multicolumn{1}{c|}{QWen-2.5 3B~\cite{xu2025qwen25omnitechnicalreport} \tiny{(baseline)}}  & 0.8531          & \multicolumn{1}{c|}{0.5314}          & -0.0059         & \multicolumn{1}{c|}{-0.0135}         & 28.20          & \multicolumn{1}{c|}{10.00}          & 51.02          \\ \hline
\multicolumn{1}{c|}{QWen-2.5 7B~\cite{xu2025qwen25omnitechnicalreport}}             & 0.8718          & \multicolumn{1}{c|}{0.4911}          & 0.0716          & \multicolumn{1}{c|}{0.0921}          & 24.60          & \multicolumn{1}{c|}{21.80}          & 57.09          \\ \hline
\multicolumn{1}{c|}{QWen-3 30BA3B~\cite{yang2025qwen3technicalreport}}           & 0.8831          & \multicolumn{1}{c|}{0.5359}          & 0.1667          & \multicolumn{1}{c|}{0.1991}          & 28.80          & \multicolumn{1}{c|}{32.20}          & 61.39          \\ \hline
\multicolumn{1}{c|}{Gemini-2.5 flash~\cite{comanici2025gemini25pushingfrontier}}        & 0.8693          & \multicolumn{1}{c|}{0.3264}          & 0.0884          & \multicolumn{1}{c|}{0.1229}          & 6.60           & \multicolumn{1}{c|}{22.60}          & 61.43          \\ \hline
\multicolumn{1}{c|}{Gemini-3 flash~\cite{gemini3flash}}          & 0.9225          & \multicolumn{1}{c|}{0.7235}          & 0.3536          & \multicolumn{1}{c|}{0.4001}          & 53.20          & \multicolumn{1}{c|}{47.80}          & 63.20          \\ \hline
\multicolumn{8}{c}{\texttt{non-LLM Methods}}                                                                                                                                                                                                    \\ \hline
\multicolumn{1}{c|}{AV-Align~\cite{AVAlign}}                & 0.8735          & \multicolumn{1}{c|}{0.4670}          & 0.1053          & \multicolumn{1}{c|}{0.1327}          & 21.00          & \multicolumn{1}{c|}{30.60}          & 55.44          \\ \hline
\multicolumn{1}{c|}{DeSync~\cite{iashin2024synchformer}}                  & 0.8830          & \multicolumn{1}{c|}{0.6160}          & 0.1355          & \multicolumn{1}{c|}{0.1587}          & 41.40          & \multicolumn{1}{c|}{15.60}          & 61.16          \\ \hline
\multicolumn{1}{c|}{JavisScore~\cite{JavisDiT}}              & 0.9284          & \multicolumn{1}{c|}{0.6747}          & 0.4268          & \multicolumn{1}{c|}{0.5041}          & 47.40          & \multicolumn{1}{c|}{50.80}          & 69.36          \\ \hline

\multicolumn{1}{c|}{RALI~\cite{zhao2025reasoning}}                    & 0.8967          & \multicolumn{1}{c|}{0.5626}          & 0.2341          & \multicolumn{1}{c|}{0.2864}          & 33.00          & \multicolumn{1}{c|}{37.80}          & 62.79          \\ \hline
\multicolumn{8}{c}{\texttt{Trained Omni-LLM}}                                                                                                                                                                                                   \\ \hline
\multicolumn{1}{c|}{Q-Align~\cite{wu2023q}}                 & 0.9163          & \multicolumn{1}{c|}{0.6791}          & 0.2927          & \multicolumn{1}{c|}{0.3633}          & 47.80          & \multicolumn{1}{c|}{32.80}          & 64.08          \\ \hline
\multicolumn{1}{c|}{Q-Insight~\cite{QInsight}}               & 0.9153          & \multicolumn{1}{c|}{0.6559}          & 0.3273          & \multicolumn{1}{c|}{0.3910}          & 46.20          & \multicolumn{1}{c|}{43.40}          & 65.78          \\ \hline
\multicolumn{1}{c|}{VQ-Insight~\cite{zhang2025vq}}              & 0.9248          & \multicolumn{1}{c|}{0.7122}          & 0.3875          & \multicolumn{1}{c|}{0.4781}          & 52.80          & \multicolumn{1}{c|}{\textbf{62.80}} & 65.85          \\ \whline
\multicolumn{1}{c|}{Ours-\tiny{w/o. R-GRPO}}        & {\ul 0.9353}    & \multicolumn{1}{c|}{{\ul 0.7316}}    & {\ul 0.4515}    & \multicolumn{1}{c|}{{\ul 0.5366}}    & {\ul 55.80}    & \multicolumn{1}{c|}{53.80}          & {\ul 71.07}    \\ \hline
\multicolumn{1}{c|}{Ours}                    & \textbf{0.9435} & \multicolumn{1}{c|}{\textbf{0.7674}} & \textbf{0.4899} & \multicolumn{1}{c|}{\textbf{0.5837}} & \textbf{61.40} & \multicolumn{1}{c|}{{\ul 55.00}}    & \textbf{72.38} \\ \whline
\end{tabular}
}}
\label{tab:main}
\end{table}

\begin{figure}[htbp]
    \centering
    
    \includegraphics[width=\linewidth]{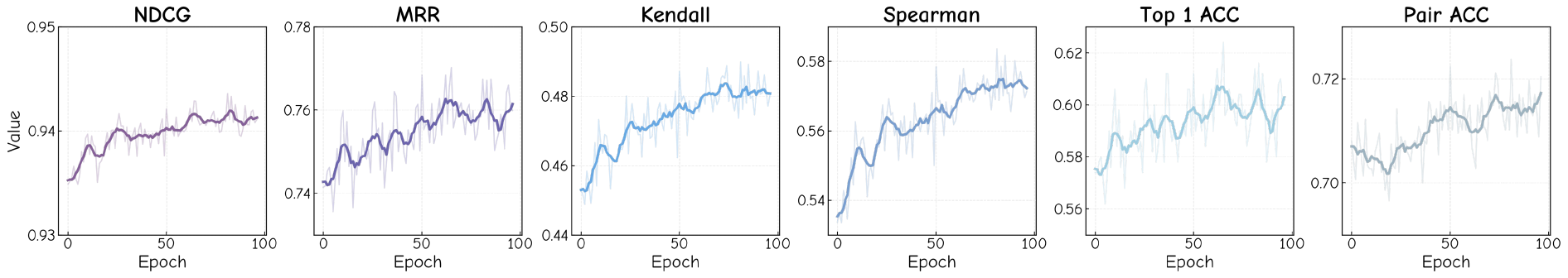}
    \caption{Learning curves on the validation set during the reinforcement learning phase. }
    \label{fig:RLrewards}
\end{figure}

As shown in Table~\ref{tab:main}, our proposed framework establishes a new state-of-the-art on the SynthSync benchmark, consistently outperforming off-the-shelf Omni-LLMs, traditional non-LLM metrics, and specifically trained LLM evaluators.
\method~yields substantial gains: direct listwise optimization boosts Top-1 Acc by 5.6\% and achieves 72.38\% Pairwise Acc. While VQ-Insight slightly leads in Bottom-1 Acc, our full \method~dominates cumulative gain and rank correlation metrics, confirming its ability to model physical causality. RL trajectories are in Fig.~\ref{fig:RLrewards}.

\vspace{-2cm}

\subsection{Benchmarking AV-Gen with SyncBench}

\noindent
\begin{minipage}[t]{0.3\linewidth}
\vspace{0pt}
  To systematically assess the causal-semantic synchronization capabilities of modern generative models, we introduce SyncBench, a comprehensive evaluation benchmark consisting of 185 diverse prompts spanning a wide array of physical events (Fig.~\ref{fig:syncbench}).
\end{minipage}%
\hfill
\begin{minipage}[t]{0.68\linewidth}
    \centering
\vspace{0pt}
    \includegraphics[width=\linewidth]{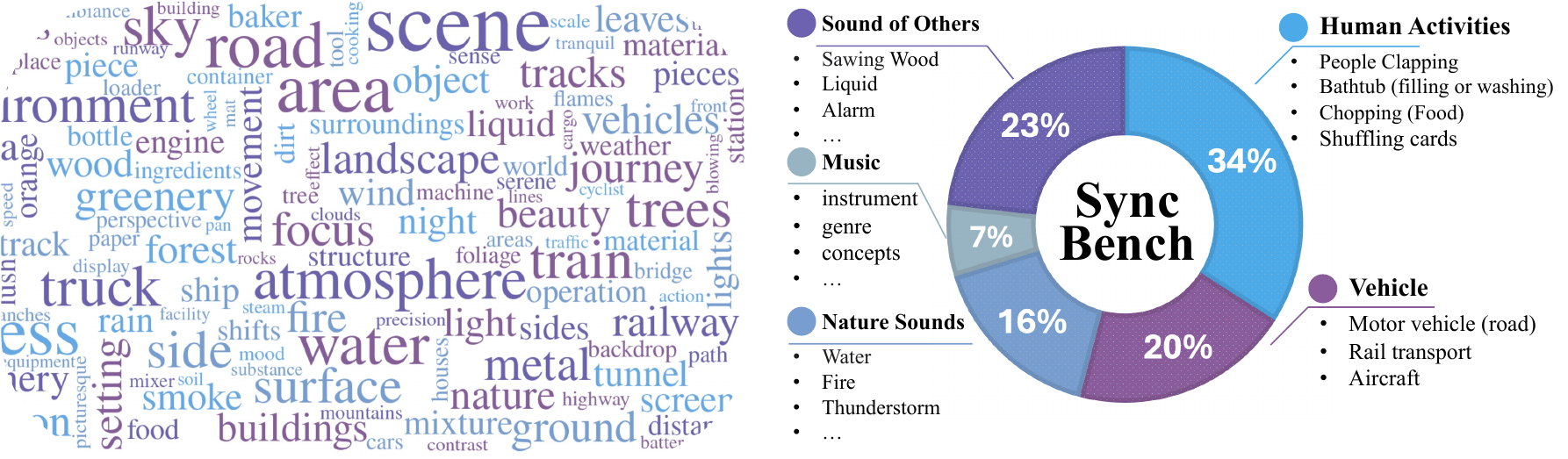}
    \captionof{figure}{Content distribution of the proposed SyncBench dataset. The left panel presents a word cloud of the 185 curated text prompts, capturing diverse physical scenarios. The right panel illustrates the categorical breakdown across five core audio-visual domains.}
    \label{fig:syncbench}
\end{minipage}

\noindent
\begin{minipage}[t]{0.3\linewidth}
\vspace{0pt}
  
 We generated audio-visual samples using six cutting-edge AV-Gen foundation models: Sora-2, Veo-3.1, WAN-2.6, Vidu-Q3, Grok-3, and LTX-2.
 As illustrated in Fig.~\ref{fig:learderboard}, our proposed metric clearly and intuitively stratifies model performance. It identifies industry-leading closed-source systems like Sora 2 and Veo 3.1 as the current state-of-the-art in physical understanding, 
\end{minipage}%
\hfill
\begin{minipage}[t]{0.68\linewidth}
    \centering
    \vspace{0pt}
    \includegraphics[width=\linewidth]{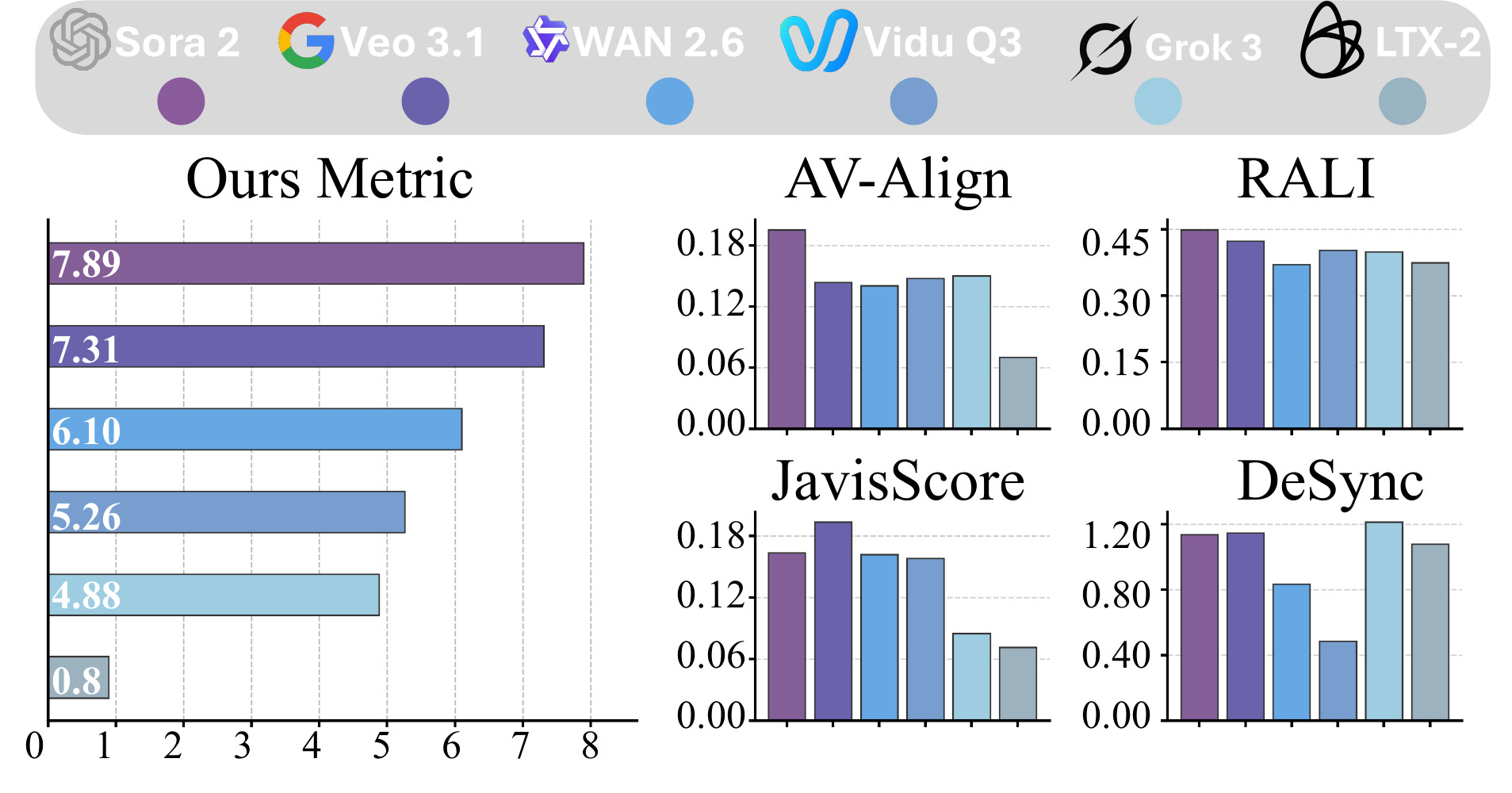}
    \captionof{figure}{Evaluation leaderboard of state-of-the-art AV-Gen models on SyncBench. The primary chart ranks the six cutting-edge models according to our proposed continuous metric. The accompanying sub-charts display the highly volatile rankings produced by traditional baseline evaluators.}
    \label{fig:learderboard}
\end{minipage}
while accurately reflecting the synchronization gap in open-weight models like LTX-2. Conversely, traditional evaluation metrics (e.g., AV-Align, JavisScore, and DeSync) exhibit extreme variance and produce rankings that often contradict human perceptual consensus. These legacy metrics frequently penalize advanced models by over-indexing on rigid, low-level signal matching rather than authentic structural causality. This stark divergence confirms that our continuous, R-GRPO-aligned metric provides a highly reliable, human-aligned standard necessary for advancing AV-Gen assessment.

\subsection{Reward Model of Test-Time Best-of-N Scaling for AV-Gen}
\noindent
\begin{minipage}[t]{0.3\linewidth}
\vspace{0pt}
We deploy our metric as an active reward for test-time Best-of-N (BoN) scaling. Using LTX-2, we generate candidates for each visual anchor and compare against existing evaluators (AV-Align, JavisScore, DeSync, RALI).
\end{minipage}%
\hfill
\begin{minipage}[t]{0.68\linewidth}
    \centering
    \vspace{0pt}
    \vspace{-5mm}
    \captionof{table}{Cross-metric generalization in BoN scaling. Rows: Reward Signal; Cols: Evaluation Metric. Diagonal (gray) excluded. Our metric yields the highest off-diagonal performance. Best valid results bolded.}
    \renewcommand{\arraystretch}{1.2}{
    \resizebox{\textwidth}{!}{
\begin{tabular}{c|c|c|c|c|c}
\whline
BoN Reward   & AV-Align                       & JavisScore                     & DeSync                         & RALI                           & Our Metric                     \\ \whline
Random       & 0.0698                         & 0.0714                         & 1.0824                         & 0.3744                         & 0.8820                         \\ \hline
+AV-Align    & \cellcolor[HTML]{EFEFEF}0.2116 & 0.0710                         & 1.0570                         & 0.3694                         & \textbf{3.2385}                \\ \hline
+ JavisScore & 0.0725                         & \cellcolor[HTML]{EFEFEF}0.1633 & \textbf{1.0280}                & 0.3901                         & 1.7417                         \\ \hline
+DeSync      & 0.0536                         & 0.0756                         & \cellcolor[HTML]{EFEFEF}0.1630 & 0.3756                         & 0.4403                         \\ \hline
+RALI        & 0.0726                         & 0.0845                         & 1.0880                         & \cellcolor[HTML]{EFEFEF}0.4596 & 1.7561                         \\ \hline
+ Ours       & \textbf{0.1439}                & \textbf{0.0934}                & 1.0336                         & \textbf{0.3959}                & \cellcolor[HTML]{EFEFEF}5.0589 \\ \whline
\end{tabular}
}}
\label{tab:ltx}
\end{minipage}
To prevent reward hacking, we strictly analyze cross-metric generalization, excluding diagonal cases where the reward and evaluation metrics match. As shown in Table~\ref{tab:ltx}, traditional metrics yield poor cross-generalization, often underperforming random selection. Conversely, our metric (+ Ours) achieves the highest off-diagonal scores across three of four independent metrics. This confirms our metric captures universal causal-semantic synchronization, proving effective for AV-Gen supervision.

\subsection{Ablation Studies}
\noindent\emph{How should an Omni-LLM parameterize rating scores for continuous evaluation?}
We validate our continuous scoring design by ablating alternative formulations. We compare \textbf{Scalar Ranking} against: 1) \textbf{Scalar Regression}, regressing ranks via MSE; 2) \textbf{Score as Text-CE}, autoregressively generating score strings; and 3) \textbf{Score as Text-KL}, minimizing KL divergence over digit probabilities.
\noindent
\begin{minipage}[t]{0.32\linewidth}
\vspace{0pt}
As Table~\ref{tab:score_rep} shows, mapping contextual judgments into rigid bounds or discrete tokens induces severe drop.
\end{minipage}%
\hfill
\begin{minipage}[t]{0.65\linewidth}
\centering
\vspace{0pt}
\vspace{-5mm}
\captionof{table}{Ablation on score representation strategies.}
\renewcommand{\arraystretch}{1.15}{
\resizebox{\textwidth}{!}{
\begin{tabular}{l >{\small}c >{\small}c >{\small}c >{\small}c}
\whline
\bfseries Score-Representation & \bfseries NDCG & \bfseries Kendall & \bfseries Pair-ACC & \bfseries Top-1 ACC \\ 
\whline
Scalar Regression    & 0.8843 & 0.1628  & 58.84    & 19.41     \\ 
Score as Text-CE     & 0.8817 & 0.1141  & 56.19    & 27.03     \\ 
Score as Text-KL     & 0.8787 & 0.1324  & 56.94    & 22.44     \\ 
\rowcolor{gray!15}
Scalar Ranking       & \textbf{0.9353} & \textbf{0.4515} & \textbf{71.07} & \textbf{55.80} \\ 
\whline
\end{tabular}
}}
\label{tab:score_rep}
\end{minipage}
 Text-based decoding and scalar regression yield poor correlation (Kendall $\le$ 0.1628) and discrimination (Top-1 Acc $<$ 28\%). Conversely, \textbf{Scalar Ranking} preserves relative topological order, boosting Pairwise Accuracy to 71.07\% and Top-1 Accuracy to 55.80\%. This confirms continuous rank optimization avoids the quantization and framing biases of discrete vocabularies.

\noindent
\begin{minipage}[t]{0.56\linewidth}
\vspace{0pt}

\noindent\emph{How does each component contribute?}
  To evaluate the individual impact of our core methodological components, we incrementally ablate the use of the SynthSync dataset training, Preference Alignment Fine-Tuning (PFT), and Reinforcement Learning (RL). The results in Table~\ref{tab:component} demonstrate the effectiveness of each component.
\end{minipage}%
\hfill
\begin{minipage}[t]{0.42\linewidth}
    \centering
    \vspace{0pt}
    \vspace{-5mm}
    \captionof{table}{Ablation study on the core components of our proposed framework.}
    \renewcommand{\arraystretch}{1.2}
    \newcommand{\imp}[1]{\textcolor{green!40!black}{\tiny #1}}
    \resizebox{\textwidth}{!}{
        \begin{tabular}{ccc >{\small}l >{\small}l} 
        \toprule
        \textbf{SynthSync} & \textbf{PFT} & \textbf{RL} & \textbf{NDCG} & \textbf{Pair ACC  (\%)} \\ 
        \midrule
        $\times$ & $\times$ & $\times$ & 0.8531 & 51.02 \\ 
        $\checkmark$ & $\times$ & $\times$ & 0.9224 \imp{(+8.1\%)} & 66.60 \imp{(+30.5\%)} \\ 
        $\checkmark$ & $\checkmark$ & $\times$ & 0.9353 \imp{(+9.6\%)} & 71.16 \imp{(+39.5\%)} \\ 
        \rowcolor{gray!15}
        $\checkmark$ & $\checkmark$ & $\checkmark$ & \textbf{0.9435} \imp{(+10.6\%)} & \textbf{72.38} \imp{(+41.9\%)} \\ 
        \bottomrule
        \end{tabular}
    }
    \label{tab:component}
\end{minipage}

\noindent
\begin{minipage}[t]{0.33\linewidth}
\vspace{0pt}
    
\noindent\emph{How do different reward formulations impact the RL post-training?}
We investigate the impact of different reward formulations during \method~post-training by comparing pairwise, listwise, and composite dual-reward 
\end{minipage}%
\hfill
\begin{minipage}[t]{0.65\linewidth}
    \centering
    \vspace{0pt}
    \vspace{-5mm}
    \captionof{table}{Ablation study of different reward functions used in the \method~framework, comparing pairwise, listwise, and dual reward strategies.}
    \renewcommand{\arraystretch}{1.2}
    \newcommand{\imp}[1]{\textcolor{green!30!black}{\tiny\,#1}}
    \resizebox{\textwidth}{!}{
    \begin{tabular}{c >{\small}l @{\hspace{0.3em}} >{\small}l @{\hspace{0.3em}} >{\small}l @{\hspace{0.3em}} >{\small}l} 
    \toprule
    \textbf{Reward} & \textbf{NDCG} & \textbf{MRR} & \textbf{Kendall} & \textbf{Pair ACC (\%)} \\ 
    \midrule
    PFT-baseline & 0.9353 & 0.7316 & 0.4515 & 71.16 \\ 
    Pair Reward  & 0.9396 \imp{(+0.5\%)} & 0.7584 \imp{(+3.7\%)} & 0.4700 \imp{(+4.1\%)} & 71.70 \imp{(+0.8\%)} \\ 
    List Reward  & 0.9339 \imp{(-0.2\%)} & 0.7371 \imp{(+0.8\%)} & 0.4267 \imp{(-5.5\%)} & 70.34 \imp{(-1.2\%)} \\ 
    \rowcolor{gray!15}
    Dual Rewards & \textbf{0.9435} \imp{(+0.9\%)} & \textbf{0.7674} \imp{(+4.9\%)} & \textbf{0.4899} \imp{(+8.5\%)} & \textbf{72.38} \imp{(+1.7\%)} \\ 
    \bottomrule
    \end{tabular}
    }
    \label{tab:rl}
\end{minipage}
strategies against the PFT baseline. As shown in Table~\ref{tab:rl}, optimizing solely with a Pair Reward yields consistent but marginal improvements across all metrics. Conversely, relying exclusively on a List Reward degrades rank correlation (Kendall's $\tau$ drops by 5.5\%) and pairwise accuracy, indicating that macroscopic distributional optimization alone can destabilize fine-grained local ordering. However, integrating both constraints into a Dual Reward strategy achieves the optimal balance. This composite approach establishes the highest overall performance, yielding a 4.9\% relative improvement in MRR and an 8.5\% improvement in Kendall's $\tau$ over the baseline. This confirms that global structural consistency and local comparative accuracy are mutually reinforcing objectives during latent continuous policy optimization.

\noindent\emph{How does the exploration variance of Gaussian policy impact RL-post training?}

\noindent
\begin{minipage}[t]{0.58\linewidth}
\vspace{0pt}
    
In our \method~ framework, the deterministic scalar output is reparameterized as the central manifold of an unconstrained exploratory Gaussian policy. Table~\ref{tab:ablate_sigma} illustrates the sensitivity of our framework to the chosen standard deviation ($\sigma$) governing this exploration space.
 When the variance is too constrained ($\sigma = 2.0$), the model lacks the necessary exploratory range to discover optimal listwise trajectories, yielding sub-optimal rank correlation (Kendall's $\tau$ of 0.4625) and pairwise accuracy (71.56\%). Conversely, an excessively large variance ($\sigma = 10.0$) introduces severe noise
\end{minipage}%
\hfill
\begin{minipage}[t]{0.4\linewidth}
    \centering
    \vspace{0pt}
    \vspace{-5mm}
    \captionof{table}{Ablation on the exploration standard deviation ($\sigma$) used in the Gaussian policy during \method. An intermediate variance yields the optimal balance between listwise exploration and policy stability, achieving peak alignment across all metrics.}
    \renewcommand{\arraystretch}{1.1}
    \newcommand{\imp}[1]{\textcolor{green!30!black}{\tiny\,#1}}
    \resizebox{\textwidth}{!}{
\begin{tabular}{cccc}
\toprule
$\sigma$  & NDCG            & Kendall         & Pair ACC       \\
\midrule
2.0  & 0.9366          & 0.4625          & 71.56          \\
\rowcolor{gray!15}
2.5  & \textbf{0.9435} & \textbf{0.4899} & \textbf{72.38} \\
10.0 & 0.9405          & 0.4768          & 71.70          \\
\bottomrule
\end{tabular}
}
\label{tab:ablate_sigma}
\end{minipage}
 into the rollout sampling, which destabilizes the proximal trust region and degrades the optimization gradient. We find that an intermediate variance of $\sigma = 2.5$ provides the optimal equilibrium---affording the policy enough flexibility to explore complex global topologies while remaining stable enough to accurately exploit learned comparative margins. This configuration achieves the peak Pairwise Accuracy of 72.38\% and an NDCG of 0.9435.

\subsection{Qualitative Visualization}
\begin{figure*}[t]
  \centering
        \includegraphics[width=0.49\linewidth]{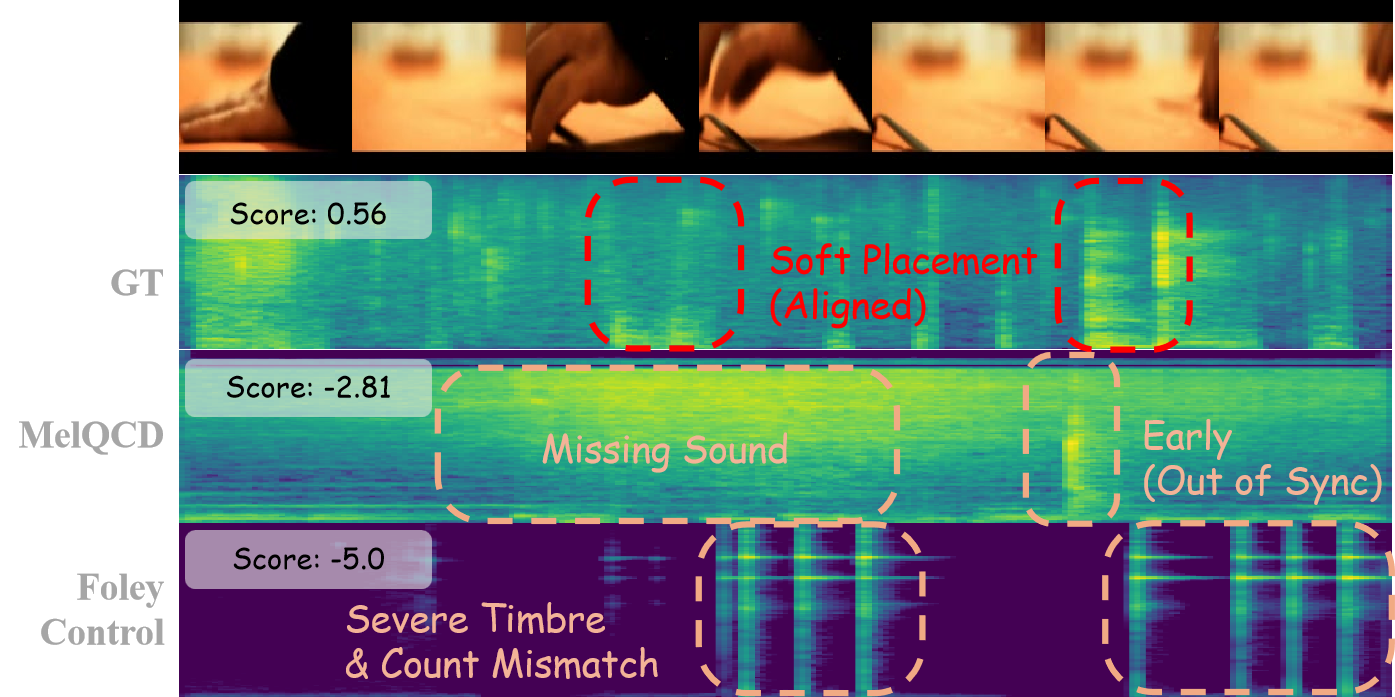}\hfill
        \includegraphics[width=0.49\linewidth]{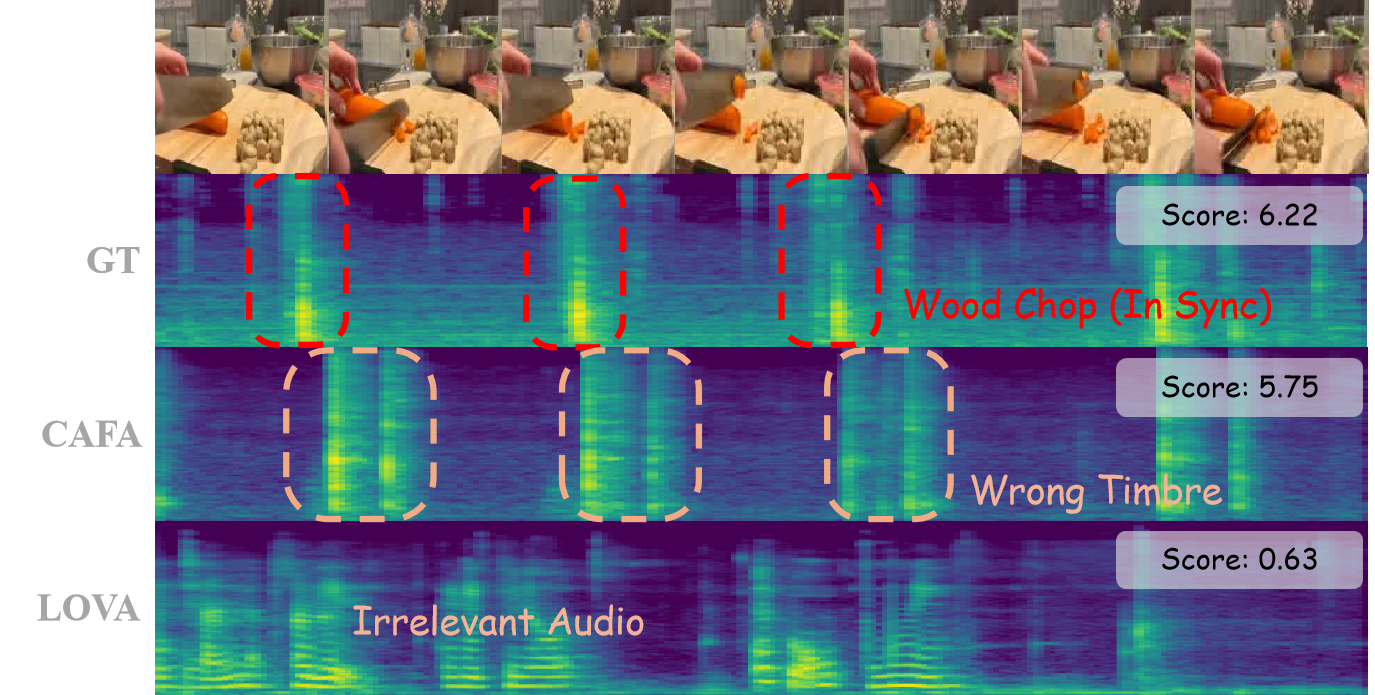}\\[0.8em]
        \includegraphics[width=0.49\linewidth]{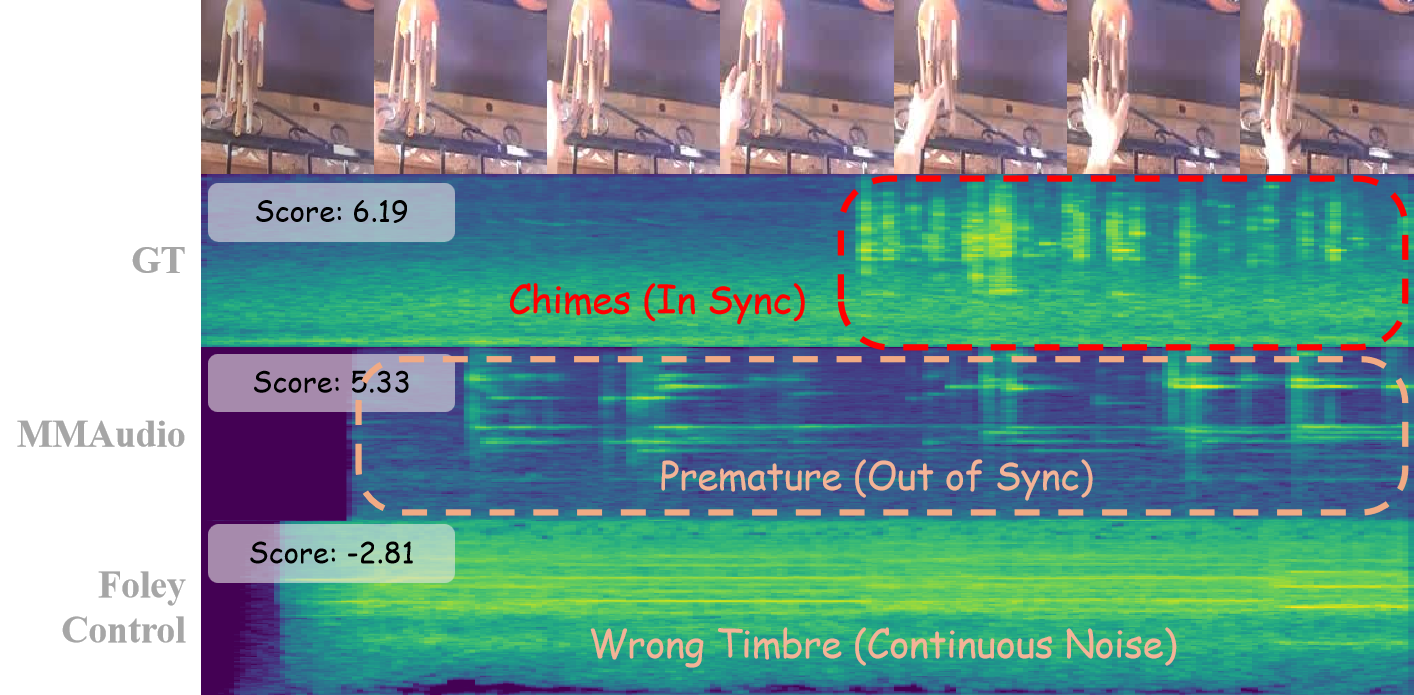}\hfill
        \includegraphics[width=0.49\linewidth]{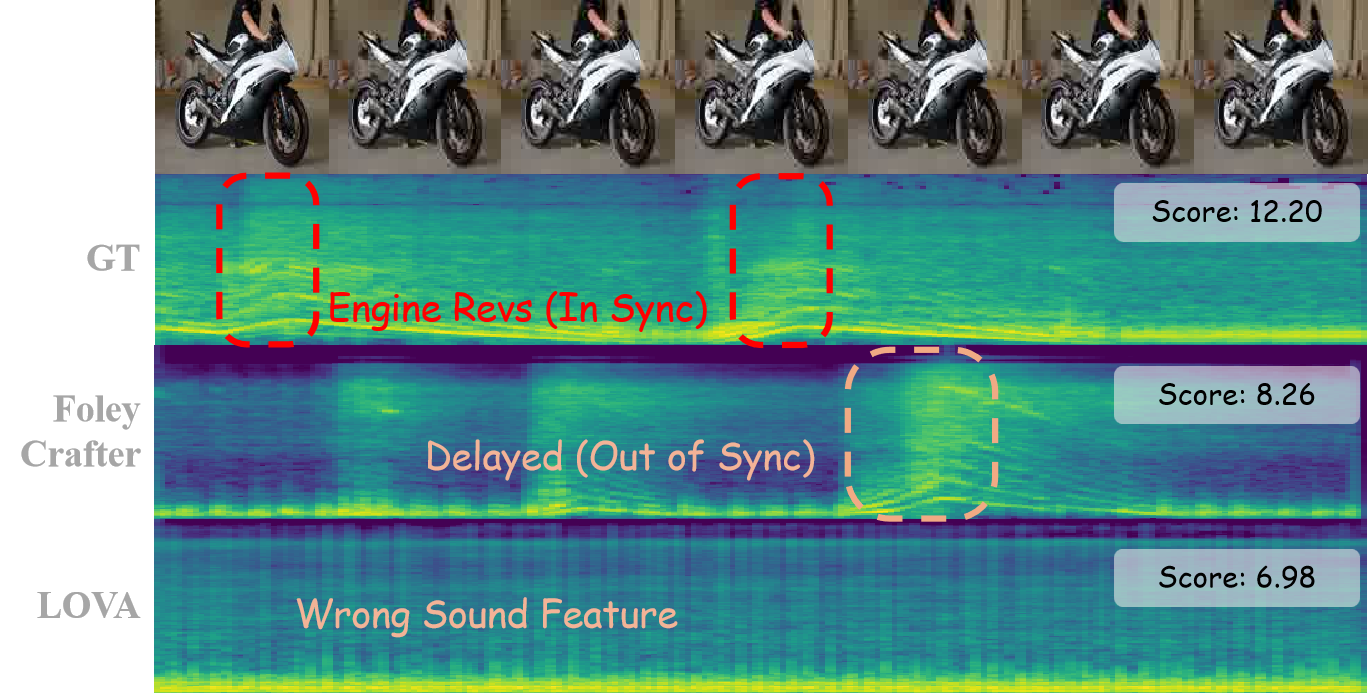}
        \caption{Qualitative visualization examples from SyncBench. \textbf{Upper left (Motorcycle Engine):} the ground truth shows clear engine-rev events aligned with the visual motion and scores highest; FoleyCrafter is delayed, while LOVA misses the acoustic signature. \textbf{Upper right (Wood Chop):} the ground truth preserves sharp, periodic chopping transients; CAFA keeps approximate timing but mismatches timbre, while LOVA is largely irrelevant. \textbf{Lower left (Soft Placement):} the ground truth contains subtle yet well-aligned contact sounds; MelQCD misses events and responds prematurely, while FoleyControl shows severe timbre distortion and event-count mismatch. \textbf{Lower right (Chimes):} the ground truth reflects synchronized chime activations; MMAudio responds prematurely, while FoleyControl is dominated by noise with incorrect timbre. Across all cases, our metric better matches causal-semantic synchronization quality than superficial signal similarity.}
  \label{fig:supp_vis_cases}
\end{figure*}

To complement the quantitative leaderboard, Fig.~\ref{fig:supp_vis_cases} shows four SyncBench examples comparing generated audio against the ground-truth pattern with the corresponding evaluator scores. Our metric consistently penalizes missing acoustic events, premature or delayed responses, timbre mismatch, and event-count inconsistency, more faithfully reflecting whether the audio preserves the causal-semantic structure implied by the visual content than low-level signal matching.

\section{Conclusion}
\label{sec:conclusion}
We identified a critical bottleneck in audio-visual generation: the lack of metrics capable of understanding intricate physical interactions between sight and sound, as traditional metrics fail on modern artifacts by assuming structural correctness and relying on simple temporal offsets. To overcome this, we present a continuous, reference-free framework for causal-semantic synchronization, built on SynthSync, a dataset of authentic generative failures ranked via human preference. By replacing the discrete language head of an Omni-LLM with a continuous latent projection and optimizing the listwise score distribution via Real-Valued Group Relative Policy Optimization (\method~), we bridge the gap between relative human perception and the need for an absolute, deployable metric. Our estimator achieves state-of-the-art human preference alignment and also generalizes robustly across metrics when deployed as a test-time Best-of-N reward model. This dual role—as both a reliable evaluator and an actionable reward signal—advances audio-visual assessment from low-level signal correlation to physically grounded causality.

\section*{Acknowledgements}
This work was supported by the Zhejiang Key Laboratory of Advanced Intelligent Warehousing and Logistics Equipment under Grant No.~2024E10007. This work was partially supported by RGC Collaborative Research Fund (No.~C5055-24G), the Start-up Fund of The Hong Kong Polytechnic University (No.~P0045999), the Seed Fund of the Research Institute for Smart Ageing (No.~P0050946), and Tsinghua-PolyU Joint Research Initiative Fund (No.~P0056509), and PolyU UGC funding (No.~P0053716).

\bibliographystyle{splncs04}
\bibliography{ref}

\begin{thebibliography}{10}
\providecommand{\url}[1]{\texttt{#1}}
\providecommand{\urlprefix}{URL }
\providecommand{\doi}[1]{https://doi.org/#1}

\bibitem{Cafa}
Benita, R., Finkelson, M., Halperin, T., Sterkin, G., Adi, Y.: Cafa: a
  controllable automatic foley artist. In: Proceedings of the IEEE/CVF
  International Conference on Computer Vision. pp. 15917--15926 (2025)

\bibitem{brokman2026training}
Brokman, J., Rachmil, O., Hofman, O., Betser, R., Giloni, A., Vainshtein, R.,
  Kojima, H.: Training-free detection of text-to-video generations via
  over-coherence. In: Proceedings of the IEEE/CVF Winter Conference on
  Applications of Computer Vision. pp. 3993--4003 (2026)

\bibitem{sora1}
Brooks, T., Peebles, B., Holmes, C., DePue, W., Guo, Y., Jing, L., Schnurr, D.,
  Taylor, J., Luhman, T., Luhman, E., Ng, C., Wang, R., Ramesh, A.: Video
  generation models as world simulators.
  \url{https://openai.com/index/video-generation-models-as-world-simulators/}
  (feb 2024), technical report. Accessed: 2026-03-01

\bibitem{seedance2}
{ByteDance Seed}: Seedance 2.0: Multi-modal ai video creation.
  \url{https://seedance2.ai/} (2026), accessed: 2026-03-01

\bibitem{cao2025multi}
Cao, H., Driouich, I., Singh, R., Thomas, E.: Multi-agent llm judge: automatic
  personalized llm judge design for evaluating natural language generation
  applications. arXiv preprint arXiv:2504.02867  (2025)

\bibitem{HunYuanFoley}
Cao, Z., Wang, T., Wang, J., Wang, Y., Zhang, Y., Chen, J., Deng, M., Wang, J.,
  Guo, Y., Liao, C., et~al.: T2av-compass: Towards unified evaluation for
  text-to-audio-video generation. arXiv preprint arXiv:2512.21094  (2025)

\bibitem{chen2020vggsoundlargescaleaudiovisualdataset}
Chen, H., Xie, W., Vedaldi, A., Zisserman, A.: Vggsound: A large-scale
  audio-visual dataset (2020), \url{https://arxiv.org/abs/2004.14368}

\bibitem{MMAudio}
Cheng, H.K., Ishii, M., Hayakawa, A., Shibuya, T., Schwing, A., Mitsufuji, Y.:
  Mmaudio: Taming multimodal joint training for high-quality video-to-audio
  synthesis. In: Proceedings of the Computer Vision and Pattern Recognition
  Conference. pp. 28901--28911 (2025)

\bibitem{Lova}
Cheng, X., Wang, X., Wu, Y., Wang, Y., Song, R.: Lova: Long-form video-to-audio
  generation. In: ICASSP 2025-2025 IEEE International Conference on Acoustics,
  Speech and Signal Processing (ICASSP). pp.~1--5. IEEE (2025)

\bibitem{cuiself}
Cui, J., Wu, J., Li, M., Yang, T., Li, X., Wang, R., Bai, A., Ban, Y., Hsieh,
  C.J.: Self-forcing++: Towards minute-scale high-quality video generation. In:
  The Fourteenth International Conference on Learning Representations

\bibitem{cui2026lol}
Cui, J., Wu, J., Li, M., Yang, T., Li, X., Wang, R., Bai, A., Ban, Y., Hsieh,
  C.J.: Lol: Longer than longer, scaling video generation to hour. arXiv
  preprint arXiv:2601.16914  (2026)

\bibitem{dixit2025foleybenchbenchmarkvideotoaudiomodels}
Dixit, S., Saito, K., Zhong, Z., Mitsufuji, Y., Donahue, C.: Foleybench: A
  benchmark for video-to-audio models (2025),
  \url{https://arxiv.org/abs/2511.13219}

\bibitem{SAO}
Evans, Z., Parker, J.D., Carr, C., Zukowski, Z., Taylor, J., Pons, J.: Stable
  audio open. In: ICASSP 2025-2025 IEEE International Conference on Acoustics,
  Speech and Signal Processing (ICASSP) (2025)

\bibitem{feng2026newtonagenticplanningphysically}
Feng, Y., Wang, J., Xu, C., Qian, Y., Wang, H., Hou, W., Liu, Y., Sun, B., Liu,
  Y., Wang, S.: Newton: Agentic planning for physically grounded video
  generation (2026), \url{https://arxiv.org/abs/2605.18396}

\bibitem{gao2025deqa}
Gao, J., Liu, R., Peng, Y., Yang, S., Zhang, J., Yang, K., You, Z.: Deqa-doc:
  Adapting deqa-score to document image quality assessment. In: Proceedings of
  the IEEE/CVF International Conference on Computer Vision. pp. 3428--3437
  (2025)

\bibitem{ImageBind}
Girdhar, R., El-Nouby, A., Liu, Z., Singh, M., Alwala, K.V., Joulin, A., Misra,
  I.: Imagebind: One embedding space to bind them all. In: Proceedings of the
  IEEE/CVF conference on computer vision and pattern recognition. pp.
  15180--15190 (2023)

\bibitem{goncalves2024perceptual}
Goncalves, L., Mathur, P., Lavania, C., Cekic, M., Federico, M., Han, K.J.:
  Perceptual evaluation of audio-visual synchrony grounded in viewers’
  opinion scores. In: European Conference on Computer Vision. pp. 288--305.
  Springer (2024)

\bibitem{veo3}
{Google DeepMind}: Veo: A text-to-video generation system.
  \url{https://deepmind.google/models/veo/} (may 2024), technical announcement.
  Accessed: 2026-03-01

\bibitem{comanici2025gemini25pushingfrontier}
{Google DeepMind}: Gemini 2.5: Pushing the frontier with advanced reasoning,
  multimodality, long context, and next generation agentic capabilities (2025),
  \url{https://arxiv.org/abs/2507.06261}

\bibitem{gemini3flash}
{Google DeepMind}: Gemini 3 flash: Frontier intelligence built for speed.
  \url{https://blog.google/products-and-platforms/products/gemini/gemini-3-flash/}
  (dec 2025), blog post. Accessed: 2026-03-01

\bibitem{guo2025deepseek}
Guo, D., Yang, D., Zhang, H., Song, J., Wang, P., Zhu, Q., Xu, R., Zhang, R.,
  Ma, S., Bi, X., et~al.: Deepseek-r1: Incentivizing reasoning capability in
  llms via reinforcement learning. arXiv preprint arXiv:2501.12948  (2025)

\bibitem{ALIVE}
Guo, Y., Gan, Q., Zhang, Y., Liu, J., Hu, Y., Xie, P., Qian, D., Zhang, Y., Li,
  R., Zhang, Y., et~al.: Alive: Animate your world with lifelike audio-video
  generation. arXiv preprint arXiv:2602.08682  (2026)

\bibitem{guo2025imagedoctor}
Guo, Y., Liu, J., Wang, Z., Chen, H., Sun, X., Zhao, Y., Wu, J., Yu, X., Liu,
  Z., Barsoum, E.: Imagedoctor: Diagnosing text-to-image generation via
  grounded image reasoning. arXiv preprint arXiv:2510.01010  (2025)

\bibitem{LTX2}
HaCohen, Y., Brazowski, B., Chiprut, N., Bitterman, Y., Kvochko, A., Berkowitz,
  A., Shalem, D., Lifschitz, D., Moshe, D., Porat, E., et~al.: Ltx-2: Efficient
  joint audio-visual foundation model. arXiv preprint arXiv:2601.03233  (2026)

\bibitem{haji2026taming}
Haji-Ali, M., Menapace, W., Siarohin, A., Balakrishnan, G., Ordonez, V.: Taming
  data and transformers for audio generation. International Journal of Computer
  Vision  \textbf{134}(3), ~87 (2026)

\bibitem{V2ALDM}
Hu, Y., Gu, Y., Li, C., Chen, R., Yu, D.: Video-to-audio generation with
  fine-grained temporal semantics. arXiv preprint arXiv:2409.14709  (2024)

\bibitem{huang2025self}
Huang, X., Li, Z., He, G., Zhou, M., Shechtman, E.: Self forcing: Bridging the
  train-test gap in autoregressive video diffusion. arXiv preprint
  arXiv:2506.08009  (2025)

\bibitem{iashin2024synchformer}
Iashin, V., Xie, W., Rahtu, E., Zisserman, A.: Synchformer: Efficient
  synchronization from sparse cues. In: ICASSP 2024-2024 IEEE International
  Conference on Acoustics, Speech and Signal Processing (ICASSP). pp.
  5325--5329. IEEE (2024)

\bibitem{jarvelin2002cumulated}
J{\"a}rvelin, K., Kek{\"a}l{\"a}inen, J.: Cumulated gain-based evaluation of ir
  techniques. ACM Transactions on Information Systems (TOIS)  \textbf{20}(4),
  422--446 (2002)

\bibitem{kendall1938new}
Kendall, M.G.: A new measure of rank correlation. Biometrika  \textbf{30}(1-2),
   81--93 (1938)

\bibitem{km2026phyeduvideo}
KM, M.M., Arun, A., Laskar, Z., Jawahar, C.: Phyeduvideo: A benchmark for
  evaluating text-to-video models for physics education. In: Proceedings of the
  IEEE/CVF Winter Conference on Applications of Computer Vision. pp. 8690--8699
  (2026)

\bibitem{Kling3}
{Kuaishou Technology}: Kling ai launches 3.0 model, ushering in an era where
  everyone can be a director (feb 2026),
  \url{https://ir.kuaishou.com/node/11216/pdf}, accessed: 2026-03-01

\bibitem{Selva}
Lee, J., Nam, J., Lee, J.: Hear what matters! text-conditioned selective
  video-to-audio generation. arXiv preprint arXiv:2512.02650  (2025)

\bibitem{ETTA}
Lee, S.g., Kong, Z., Goel, A., Kim, S., Valle, R., Catanzaro, B.: Etta:
  Elucidating the design space of text-to-audio models. In: International
  Conference on Machine Learning (2025)

\bibitem{OmniCustom}
Li, M., Li, Z., Zhang, K., Yin, G., Li, Z., Xu, D.: Omnicustom: Sync
  audio-video customization via joint audio-video generation model. arXiv
  preprint arXiv:2602.12304  (2026)

\bibitem{QInsight}
Li, W., Zhang, X., Zhao, S., Zhang, Y., Li, J., Zhang, L., Zhang, J.:
  Q-insight: Understanding image quality via visual reinforcement learning.
  arXiv preprint arXiv:2503.22679  (2025)

\bibitem{Thinksound}
Liu, H., Luo, K., Wang, J., Wang, W., Chen, Q., Zhao, Z., Xue, W.: Thinksound:
  Chain-of-thought reasoning in multimodal large language models for audio
  generation and editing. arXiv preprint arXiv:2506.21448  (2025)

\bibitem{PrismAudio}
Liu, H., Luo, K., Wang, W., Chen, Q., Sun, P., Huang, R., Li, X., Ye, J., Xue,
  W.: Prismaudio: Decomposed chain-of-thoughts and multi-dimensional rewards
  for video-to-audio generation. arXiv preprint arXiv:2511.18833  (2025)

\bibitem{JavisDiT}
Liu, K., Li, W., Chen, L., Wu, S., Zheng, Y., Ji, J., Zhou, F., Jiang, R., Luo,
  J., Fei, H., et~al.: Javisdit: Joint audio-video diffusion transformer with
  hierarchical spatio-temporal prior synchronization. arXiv preprint
  arXiv:2503.23377  (2025)

\bibitem{javisdit++}
Liu, K., Zheng, Y., Wang, K., Wu, S., Zhang, R., Luo, J., Hatzinakos, D., Liu,
  Z., Fei, H., Chua, T.S.: Javisdit++: Unified modeling and optimization for
  joint audio-video generation. arXiv preprint arXiv:2602.19163  (2026)

\bibitem{Ma_2026_CVPR}
Ma, Y., Wang, X., Ma, Q., Wang, Q., Zheng, M., Yang, X., Li, H., Zhao, C.,
  Ying, J., Yang, H., Liu, H., Chen, Q.: Group editing: Edit multiple images in
  one go. In: Proceedings of the IEEE/CVF Conference on Computer Vision and
  Pattern Recognition (CVPR). pp. 43418--43428 (June 2026)

\bibitem{ma2026fastvmteliminatingredundancy}
Ma, Y., Wang, Z., Ren, T., Zheng, M., Liu, H., Guo, J., Feng, K., Xue, Y.,
  Zhao, Z., Schindler, K., Chen, Q., Zhang, L.: Fastvmt: Eliminating redundancy
  in video motion transfer (2026), \url{https://arxiv.org/abs/2602.05551}

\bibitem{ma2026easyvfxfrequencydrivendecoupling}
Ma, Y., Ye, X., Wang, Q., Wang, Y., Liu, H., Zhang, Y., Wang, X., Che, Y., Mo,
  S., Liang, P., Zhan, F., Chen, Q.: Easyvfx: Frequency-driven decoupling for
  resource-efficient vfx generation (2026),
  \url{https://arxiv.org/abs/2605.22051}

\bibitem{mohammadkhani2025checklist}
Mohammadkhani, M.G., Beigy, H.: Checklist engineering empowers multilingual llm
  judges. In: Proceedings of the Workshop on Beyond English: Natural Language
  Processing for all Languages in an Era of Large Language Models. pp. 190--196
  (2025)

\bibitem{mu2025evaluate}
Mu, W., Xu, L., Pei, S., Mi, L., Zhou, H.: Evaluate-and-purify: Fortifying code
  language models against adversarial attacks using llm-as-a-judge. In:
  International Conference on Intelligent Computing. pp. 369--379. Springer
  (2025)

\bibitem{veo2}
van~den Oord, A., Roman, E.: State-of-the-art video and image generation with
  veo 2 and imagen 3 (dec 2024),
  \url{https://blog.google/innovation-and-ai/models-and-research/google-labs/video-image-generation-update-december-2024/},
  accessed: 2026-03-01

\bibitem{openai2025sora2}
{OpenAI}: Sora 2 is here. \url{https://openai.com/index/sora-2/} (sep 2025),
  blog post. Accessed: 2026-03-01

\bibitem{ouyang2022training}
Ouyang, L., Wu, J., Jiang, X., Almeida, D., Wainwright, C., Mishkin, P., Zhang,
  C., Agarwal, S., Slama, K., Ray, A., et~al.: Training language models to
  follow instructions with human feedback. Advances in neural information
  processing systems  \textbf{35},  27730--27744 (2022)

\bibitem{qian2025thinkmovelatentmotion}
Qian, Y., Wang, J., Feng, Y., Xu, C., Lu, W., Liu, Y., Sun, B., Chen, Y., Liu,
  Y., Wang, S.: Think before you move: Latent motion reasoning for
  text-to-motion generation (2025), \url{https://arxiv.org/abs/2512.24100}

\bibitem{rafailov2023direct}
Rafailov, R., Sharma, A., Mitchell, E., Manning, C.D., Ermon, S., Finn, C.:
  Direct preference optimization: Your language model is secretly a reward
  model. Advances in Neural Information Processing Systems  \textbf{36},
  53728--53741 (2023)

\bibitem{Syncnet}
Raina, A., Arora, V.: Syncnet: Correlating objective for time delay estimation
  in audio signals. In: ICASSP 2023-2023 IEEE International Conference on
  Acoustics, Speech and Signal Processing (ICASSP). pp.~1--5. IEEE (2023)

\bibitem{rowles2025foley}
Rowles, C., Jampani, V., Donn{\'e}, S., Vainer, S., Parker, J., Evans, Z.:
  Foley control: Aligning a frozen latent text-to-audio model to video. arXiv
  preprint arXiv:2510.21581  (2025)

\bibitem{SeD15P}
Seedance, T., Chen, H., Chen, S., Chen, X., Chen, Y., Chen, Y., Chen, Z.,
  Cheng, F., Cheng, T., Cheng, X., et~al.: Seedance 1.5 pro: A native
  audio-visual joint generation foundation model. arXiv preprint
  arXiv:2512.13507  (2025)

\bibitem{shibata2025lces}
Shibata, T., Miyamura, Y.: Lces: Zero-shot automated essay scoring via pairwise
  comparisons using large language models. In: Proceedings of the 2025
  Conference on Empirical Methods in Natural Language Processing. pp.
  29976--29989 (2025)

\bibitem{song2025syncphony}
Song, J., Kwon, M., Jeong, J., Uh, Y.: Syncphony: Synchronized audio-to-video
  generation with diffusion transformers. arXiv preprint arXiv:2509.21893
  (2025)

\bibitem{MOVA}
Team, O., Yu, D., Chen, M., Chen, Q., Luo, Q., Wu, Q., Cheng, Q., Li, R.,
  Liang, T., Zhang, W., et~al.: Mova: Towards scalable and synchronized
  video-audio generation. arXiv preprint arXiv:2602.08794  (2026)

\bibitem{AudioX}
Tian, Z., Jin, Y., Liu, Z., Yuan, R., Tan, X., Chen, Q., Xue, W., Guo, Y.:
  Audiox: Diffusion transformer for anything-to-audio generation. arXiv
  preprint arXiv:2503.10522  (2025)

\bibitem{tutz1986bradley}
Tutz, G.: Bradley-terry-luce models with an ordered response. Journal of
  mathematical psychology  \textbf{30}(3),  306--316 (1986)

\bibitem{Fugatto}
Valle, R., Badlani, R., Kong, Z., Lee, S.g., Goel, A., Kim, S., Santos, J.F.,
  Dai, S., Gururani, S., Aljafari, A., et~al.: Fugatto 1: Foundational
  generative audio transformer opus 1. In: International Conference on Learning
  Representations (2025)

\bibitem{wan2025wan}
Wan, T., Wang, A., Ai, B., Wen, B., Mao, C., Xie, C.W., Chen, D., Yu, F., Zhao,
  H., Yang, J., et~al.: Wan: Open and advanced large-scale video generative
  models. arXiv preprint arXiv:2503.20314  (2025)

\bibitem{UniVerse}
Wang, D., Zuo, W., Li, A., Chen, L.H., Liao, X., Zhou, D., Yin, Z., Dai, X.,
  Jiang, D., Yu, G.: Universe-1: Unified audio-video generation via stitching
  of experts. arXiv preprint arXiv:2509.06155  (2025)

\bibitem{Alignnet}
Wang, J., Fang, Z., Zhao, H.: Alignnet: A unifying approach to audio-visual
  alignment. In: Proceedings of the IEEE/CVF Winter Conference on Applications
  of Computer Vision. pp. 3309--3317 (2020)

\bibitem{Klear}
Wang, J., Qiang, C., Guo, Y., Wang, Y., Zeng, X., Zhang, C., Wan, P.: Klear:
  Unified multi-task audio-video joint generation. arXiv preprint
  arXiv:2601.04151  (2026)

\bibitem{wang2026guided}
Wang, J., Hu, Z., Xu, C., Ren, S., Feng, Y., Liu, Y., Sun, B., Wang, S.: Guided
  by the plan: Enhancing faithful autoregressive text-to-audio generation with
  guided decoding. arXiv preprint arXiv:2601.14304  (2026)

\bibitem{wang2025language}
Wang, J., Xu, C., Yu, C., Hu, Z., Xie, H., Yu, G., Shang, L., Wang, S.:
  Language model based text-to-audio generation: Anti-causally aligned
  collaborative residual transformers. In: Proceedings of the 2025 Conference
  on Empirical Methods in Natural Language Processing. pp. 26036--26054 (2025)

\bibitem{MelQCD}
Wang, J., Xu, C., Yu, C., Shang, L., Hu, Z., Wang, S., Bo, L.: Synchronized
  video-to-audio generation via mel quantization-continuum decomposition. In:
  Proceedings of the Computer Vision and Pattern Recognition Conference. pp.
  3111--3120 (2025)

\bibitem{wang2025contrastscore}
Wang, X., Larionov, D., Wu, S., Liu, Y., Eger, S., Moosavi, N.S., Lin, C.:
  Contrastscore: Towards higher quality, less biased, more efficient evaluation
  metrics with contrastive evaluation. In: Proceedings of the 14th
  International Joint Conference on Natural Language Processing and the 4th
  Conference of the Asia-Pacific Chapter of the Association for Computational
  Linguistics. pp. 3045--3060 (2025)

\bibitem{weng2025audiosyncvideogenerationmultistream}
Weng, S., Zheng, H., Chang, Z., Li, S., Shi, B., Wang, X.: Audio-sync video
  generation with multi-stream temporal control (2025),
  \url{https://arxiv.org/abs/2506.08003}

\bibitem{whitehouse2025j1}
Whitehouse, C., Wang, T., Yu, P., Li, X., Weston, J., Kulikov, I., Saha, S.:
  J1: Incentivizing thinking in llm-as-a-judge via reinforcement learning.
  arXiv preprint arXiv:2505.10320  (2025)

\bibitem{wu2023q}
Wu, H., Zhang, Z., Zhang, W., Chen, C., Liao, L., Li, C., Gao, Y., Wang, A.,
  Zhang, E., Sun, W., et~al.: Q-align: Teaching lmms for visual scoring via
  discrete text-defined levels. arXiv preprint arXiv:2312.17090  (2023)

\bibitem{wu2025visualquality}
Wu, T., Zou, J., Liang, J., Zhang, L., Ma, K.: Visualquality-r1:
  Reasoning-induced image quality assessment via reinforcement learning to
  rank. arXiv preprint arXiv:2505.14460  (2025)

\bibitem{xu2024facechain}
Xu, C., Liu, Y., Xing, J., Wang, W., Sun, M., Dan, J., Huang, T., Li, S.,
  Cheng, Z.Q., Tai, Y., et~al.: Facechain-imagineid: Freely crafting
  high-fidelity diverse talking faces from disentangled audio. In: Proceedings
  of the IEEE/CVF Conference on Computer Vision and Pattern Recognition. pp.
  1292--1302 (2024)

\bibitem{xu2023high}
Xu, C., Zhu, J., Zhang, J., Han, Y., Chu, W., Tai, Y., Wang, C., Xie, Z., Liu,
  Y.: High-fidelity generalized emotional talking face generation with
  multi-modal emotion space learning. In: Proceedings of the IEEE/CVF
  conference on computer vision and pattern recognition. pp. 6609--6619 (2023)

\bibitem{xu2025qwen25omnitechnicalreport}
Xu, J., Guo, Z., He, J., Hu, H., He, T., Bai, S., Chen, K., Wang, J., Fan, Y.,
  Dang, K., Zhang, B., Wang, X., Chu, Y., Lin, J.: Qwen2.5-omni technical
  report (2025), \url{https://arxiv.org/abs/2503.20215}

\bibitem{yang2025qwen3technicalreport}
Yang, A., Li, A., Yang, B., Zhang, B., Hui, B., Zheng, B., Yu, B., Gao, C.,
  Huang, C., Lv, C., Zheng, C., Liu, D., Zhou, F., Huang, F., Hu, F., Ge, H.,
  Wei, H., Lin, H., Tang, J., Yang, J., Tu, J., Zhang, J., Yang, J., Yang, J.,
  Zhou, J., Zhou, J., Lin, J., Dang, K., Bao, K., Yang, K., Yu, L., Deng, L.,
  Li, M., Xue, M., Li, M., Zhang, P., Wang, P., Zhu, Q., Men, R., Gao, R., Liu,
  S., Luo, S., Li, T., Tang, T., Yin, W., Ren, X., Wang, X., Zhang, X., Ren,
  X., Fan, Y., Su, Y., Zhang, Y., Zhang, Y., Wan, Y., Liu, Y., Wang, Z., Cui,
  Z., Zhang, Z., Zhou, Z., Qiu, Z.: Qwen3 technical report (2025),
  \url{https://arxiv.org/abs/2505.09388}

\bibitem{AVAlign}
Yariv, G., Gat, I., Benaim, S., Wolf, L., Schwartz, I., Adi, Y.: Diverse and
  aligned audio-to-video generation via text-to-video model adaptation. In:
  Proceedings of the AAAI Conference on Artificial Intelligence. vol.~38, pp.
  6639--6647 (2024)

\bibitem{zhang2026reward}
Zhang, J., Li, N., Ban, Y., Bai, A., Cui, J.: Reward-forcing: Autoregressive
  video generation with reward feedback. arXiv preprint arXiv:2601.16933
  (2026)

\bibitem{zhang2025agent}
Zhang, S., Yin, M., Zhang, J., Liu, J., Han, Z., Zhang, J., Li, B., Wang, C.,
  Wang, H., Chen, Y., et~al.: Which agent causes task failures and when? on
  automated failure attribution of llm multi-agent systems. arXiv preprint
  arXiv:2505.00212  (2025)

\bibitem{zhang2025vq}
Zhang, X., Li, W., Zhao, S., Li, J., Zhang, L., Zhang, J.: Vq-insight: Teaching
  vlms for ai-generated video quality understanding via progressive visual
  reinforcement learning. arXiv preprint arXiv:2506.18564  (2025)

\bibitem{FoleyCrafter}
Zhang, Y., Gu, Y., Zeng, Y., Xing, Z., Wang, Y., Wu, Z., Liu, B., Chen, K.:
  Foleycrafter: Bring silent videos to life with lifelike and synchronized
  sounds. International Journal of Computer Vision  \textbf{134}(1), ~46 (2026)

\bibitem{zhao2025reasoning}
Zhao, S., Zhang, X., Li, W., Li, J., Zhang, L., Xue, T., Zhang, J.: Reasoning
  as representation: Rethinking visual reinforcement learning in image quality
  assessment. arXiv preprint arXiv:2510.11369  (2025)

\bibitem{zheng2026aligning}
Zheng, J., Pan, S., Yao, Y., Wang, Z., Wang, D., Liu, T.: Aligning what
  matters: Masked latent adaptation for text-to-audio-video generation.
  Advances in Neural Information Processing Systems  \textbf{38},
  173244--173272 (2026)

\bibitem{zhu2024adaptive}
Zhu, H., Wu, H., Li, Y., Zhang, Z., Chen, B., Zhu, L., Fang, Y., Zhai, G., Lin,
  W., Wang, S.: Adaptive image quality assessment via teaching large multimodal
  model to compare. Advances in Neural Information Processing Systems
  \textbf{37},  32611--32629 (2024)

\bibitem{zhu2026causal}
Zhu, H., Zhao, M., He, G., Su, H., Li, C., Zhu, J.: Causal forcing:
  Autoregressive diffusion distillation done right for high-quality real-time
  interactive video generation. arXiv preprint arXiv:2602.02214  (2026)

\end{thebibliography}
\clearpage
\appendix
\setcounter{page}{1}
\section{Appendix}
\label{sec:appendix}

\subsection{Dataset Statistics and Split}

\noindent\emph{Dataset Sources and Composition.} Our SynthSync dataset is curated from two complementary video corpora: VGGSound\cite{chen2020vggsoundlargescaleaudiovisualdataset} and FoleyBench\cite{dixit2025foleybenchbenchmarkvideotoaudiomodels}. After temporal standardization to 5-second clips, the final dataset contains 2,267 visual anchors in total. Among them, 1,016 samples originate from VGGSound, while the remaining 1,251 samples are drawn from FoleyBench. The source origins are systematically tracked throughout preprocessing and annotation to ensure balanced representation and transparent evaluation.

\noindent\emph{Data Split.} To prevent content leakage and ensure a strict evaluation protocol, we partition the dataset at the video level rather than at the generated-audio level. Specifically, all candidate audio generations associated with the same visual anchor are assigned to the same split. The resulting partition contains 1,767 training samples and 500 test samples. This design prevents near-duplicate visual content from appearing across training and evaluation, ensuring that the reported performance reflects genuine generalization in causal-semantic synchronization assessment.

\subsection{Best-of-N Selection and Cross-Metric Evaluation Protocol}
\noindent\emph{Candidate Generation Protocol.} For each prompt in SyncBench, we construct a fixed candidate pool under the same textual condition using the LTX-2 generator. Specifically, we first sample one batch of 6 videos using random seed 42, and then repeat the same generation procedure with random seed 1234 to obtain another batch of 6 videos. Therefore, each prompt is associated with 12 candidate videos in total. This two-seed design increases sample diversity while keeping the generation budget and prompt condition strictly controlled across all reward models.

\noindent\emph{Best-of-N Selection.} We then evaluate every candidate video under all five scoring schemes, namely AV-Align, JavisScore, DeSync, RALI, and our proposed metric. For a given reward model, BoN selection is performed only within the 12 candidates corresponding to the same prompt: we identify the single candidate that achieves the highest score under that reward model and mark it as the selected sample for that prompt. Importantly, after this selection is made, we record not only the reward score used for selection itself, but also the scores assigned to the same selected sample by the other four evaluators. Repeating this procedure over all prompts yields one cross-metric performance profile for each reward model.

\noindent\emph{Cross-Metric Generalization.} This protocol is deliberately designed to test generalization rather than self-consistency. A reward model is not judged solely by whether it can select samples that maximize its own score, since such behavior can arise from reward overfitting or metric-specific bias. Instead, a reward model is considered effective only if the samples it selects are also rated favorably by other independent evaluators. Consequently, the off-diagonal entries in Table~\ref{tab:ltx} constitute the principal evidence for reward quality. Strong off-diagonal performance indicates that the reward captures transferable causal-semantic synchronization, whereas weak transfer suggests the reward is overly tied to local signal artifacts or narrow objective shortcuts.

\subsection{Comprehensive Ablation Studies}
\noindent\emph{Extended Ablations.} We include the complete ablation results in the appendix to complement the compact presentation in the main paper. In addition to the core component analysis, the appendix consolidates the full reward-function comparison, Gaussian exploration variance study, and any supplementary ablations omitted from the main paper for space considerations. This expanded presentation clarifies the individual contribution of dataset curation, preference-aligned initialization, and reinforcement learning to the final synchronization estimator.

\noindent\emph{Complete Tables.} Tables~\ref{tab:score_rep_full}--\ref{tab:ablate_sigma_full} expand the ablations in the main paper with the full evaluation suite. Table~\ref{tab:score_rep_full} shows that scalar ranking remains the strongest score parameterization across all ranking and decision metrics. Table~\ref{tab:module_full} presents the cumulative contribution of SynthSync training, preference-aligned fine-tuning, and RL post-training, showing consistent gains in both global ranking quality and top/bottom decision accuracy. Table~\ref{tab:rl_full} further confirms that the dual-reward design provides the best overall trade-off, whereas a list-only reward weakens several metrics despite preserving reasonable cumulative gain. Finally, Table~\ref{tab:ablate_sigma_full} indicates that an intermediate exploration variance of $\sigma=2.5$ achieves the best overall balance, while a larger variance slightly improves Last-1 accuracy at the cost of weaker overall alignment.

\begin{table*}[t]
\centering
\caption{Complete ablation on score representation strategies. This table extends the compact comparison in the main paper with the full suite of ranking and decision metrics.}
\label{tab:score_rep_full}
\renewcommand{\arraystretch}{1.12}
\resizebox{\textwidth}{!}{
\begin{tabular}{lccccccc}
\whline
\textbf{Score Representation} & \textbf{NDCG} & \textbf{MRR} & \textbf{Kendall} & \textbf{Spearman} & \textbf{Top-1 ACC} & \textbf{Pair ACC} & \textbf{Last ACC} \\ \whline
Scalar Regression & 0.8843 & 0.4656 & 0.1628 & 0.2088 & 19.4 & 58.84 & 35.2 \\
Score as Text-CE & 0.8736 & 0.4442 & 0.0859 & 0.1033 & 20.2 & 56.46 & 28.0 \\
Score as Text-KL & 0.8787 & 0.4626 & 0.1063 & 0.1324 & 22.4 & 56.94 & 29.8 \\
\rowcolor{gray!15}
Scalar Ranking (Ours) & \textbf{0.9353} & \textbf{0.7316} & \textbf{0.4515} & \textbf{0.5366} & \textbf{55.8} & \textbf{71.07} & \textbf{53.8} \\
\whline
\end{tabular}}
\end{table*}

\begin{table*}[t]
\centering
\caption{Complete module ablation aligned with the SynthSync/PFT/RL component study in the main paper. We report the full metric suite for each cumulative training stage.}
\label{tab:module_full}
\renewcommand{\arraystretch}{1.2}
\newcommand{\pos}[1]{\textcolor{green!40!black}{\tiny #1}}
\resizebox{\textwidth}{!}{
\begin{tabular}{ccc >{\small}l >{\small}l >{\small}l >{\small}l >{\small}l >{\small}l >{\small}l}
\toprule
\textbf{SynthSync} & \textbf{PFT} & \textbf{RL} & \textbf{NDCG} & \textbf{MRR} & \textbf{Kendall} & \textbf{Spearman} & \textbf{Top-1 ACC} & \textbf{Pair ACC} & \textbf{Last ACC} \\
\midrule
$\times$ & $\times$ & $\times$ & 0.8531 & 0.5314 & -0.0059 & -0.0135 & 28.20 & 51.02 & 10.0 \\
$\checkmark$ & $\times$ & $\times$ & 0.9224 \pos{(+8.1\%)} & 0.6554 \pos{(+23.3\%)} & 0.3708 \pos{(+0.3767)} & 0.4559 \pos{(+0.4694)} & 44.2 \pos{(+56.7\%)} & 66.60 \pos{(+30.5\%)} & 48.6 \pos{(+386.0\%)} \\
$\checkmark$ & $\checkmark$ & $\times$ & 0.9353 \pos{(+9.6\%)} & 0.7316 \pos{(+37.7\%)} & 0.4515 \pos{(+0.4574)} & 0.5366 \pos{(+0.5501)} & 55.80 \pos{(+97.9\%)} & 71.16 \pos{(+39.5\%)} & 53.8 \pos{(+438.0\%)} \\
\rowcolor{gray!15}
$\checkmark$ & $\checkmark$ & $\checkmark$ & \textbf{0.9435} \pos{(+10.6\%)} & \textbf{0.7674} \pos{(+44.4\%)} & \textbf{0.4899} \pos{(+0.4958)} & \textbf{0.5837} \pos{(+0.5972)} & \textbf{61.4} \pos{(+117.7\%)} & \textbf{72.38} \pos{(+41.9\%)} & \textbf{55.0} \pos{(+450.0\%)} \\
\bottomrule
\end{tabular}}
\end{table*}

\begin{table*}[t]
\centering
\caption{Complete reward ablation aligned with the pairwise, listwise, and dual-reward comparison in the main paper. We report the full metric suite together with relative changes against the PFT baseline.}
\label{tab:rl_full}
\renewcommand{\arraystretch}{1.2}
\newcommand{\pos}[1]{\textcolor{green!30!black}{\tiny\,#1}}
\resizebox{\textwidth}{!}{
\begin{tabular}{lccccccc}
\toprule
\textbf{Reward} & \textbf{NDCG} & \textbf{MRR} & \textbf{Kendall} & \textbf{Spearman} & \textbf{Top-1 ACC} & \textbf{Pair ACC} & \textbf{Last ACC} \\
\midrule
PFT-baseline & 0.9353 & 0.7316 & 0.4515 & 0.5366 & 55.80 & 71.16 & 53.8 \\
Pair Reward & 0.9396 \pos{(+0.5\%)} & 0.7584 \pos{(+3.7\%)} & 0.4700 \pos{(+4.1\%)} & 0.5560 \pos{(+3.6\%)} & 60.00 \pos{(+7.5\%)} & 71.70 \pos{(+0.8\%)} & 56.4 \pos{(+4.8\%)} \\
List Reward & 0.9339 \pos{(-0.2\%)} & 0.7371 \pos{(+0.8\%)} & 0.4267 \pos{(-5.5\%)} & 0.5122 \pos{(-4.5\%)} & 55.6 \pos{(-0.4\%)} & 70.34 \pos{(-1.2\%)} & 44.0 \pos{(-18.2\%)} \\
\rowcolor{gray!15}
Dual Rewards & \textbf{0.9435} \pos{(+0.9\%)} & \textbf{0.7674} \pos{(+4.9\%)} & \textbf{0.4899} \pos{(+8.5\%)} & \textbf{0.5837} \pos{(+8.8\%)} & \textbf{61.4} \pos{(+10.0\%)} & \textbf{72.38} \pos{(+1.7\%)} & \textbf{55.0} \pos{(+2.2\%)} \\
\bottomrule
\end{tabular}}
\end{table*}

\begin{table*}[t]
\centering
\caption{Complete ablation on the Gaussian exploration standard deviation ($\sigma$) used during RL post-training. The intermediate setting $\sigma=2.5$ gives the strongest overall trade-off, while $\sigma=10.0$ yields the best Last ACC.}
\label{tab:ablate_sigma_full}
\renewcommand{\arraystretch}{1.1}
\resizebox{\textwidth}{!}{
\begin{tabular}{cccccccc}
\whline
$\sigma$ & NDCG & MRR & Kendall & Spearman & Top-1 ACC & Pair ACC & Last ACC \\ \whline
2.0 & 0.9366 & 0.7327 & 0.4625 & 0.5493 & 56.20 & 71.56 & 54.00 \\
\rowcolor[HTML]{EFEFEF}
2.5 & \textbf{0.9435} & \textbf{0.7674} & \textbf{0.4899} & \textbf{0.5837} & \textbf{61.40} & \textbf{72.38} & 55.0 \\
10.0 & 0.9405 & 0.7503 & 0.4768 & 0.5673 & 58.60 & 71.70 & \textbf{55.60} \\
\whline
\end{tabular}}
\end{table*}

\subsection{Global Reward and Metric Formulations}
\noindent\emph{Evaluation Metrics.} For each visual anchor $x$, let $\mathcal{Y}_{x}=\{y_{1},\dots,y_{K}\}$ denote the candidate audio-video pairs, with predicted scores $\hat{\mathbf{s}}=[\hat{s}_{1},\dots,\hat{s}_{K}]$ and ground-truth scores $\mathbf{g}=[g_{1},\dots,g_{K}]$. The predicted ranking is obtained by sorting candidates in descending order of $\hat{\mathbf{s}}$, while the ground-truth ranking is induced by descending order of $\mathbf{g}$. All ranking metrics are computed from these ordered lists.

\noindent\emph{Pairwise Accuracy.} Given a valid pair $(i,j)$ with $g_i \neq g_j$, PairAcc evaluates whether the predicted ordering agrees with the ground-truth ordering:
\[
\mathrm{PairAcc}
=
\frac{1}{|\mathcal{P}|}
\sum_{(i,j)\in\mathcal{P}}
\mathbb{I}\!\left[
\left(\hat{s}_{i}>\hat{s}_{j}\right)
\Leftrightarrow
\left(g_{i}>g_{j}\right)
\right],
\]
where $\mathcal{P}$ is the set of annotated comparison pairs. Pairs with tied ground-truth scores are skipped.

\noindent\emph{NDCG.} Let $\pi$ be the predicted permutation and let the relevance at rank $r$ be defined by the ground-truth score of the item placed at that position, namely $\mathrm{rel}_{r}=g_{\pi(r)}$. The discounted cumulative gain is
\[
\mathrm{DCG}(\pi)=\sum_{r=1}^{K}\frac{\mathrm{rel}_{r}}{\log_{2}(r+1)},
\]
and the normalized metric is
\[
\mathrm{NDCG}
=
\frac{\mathrm{DCG}(\pi)}{\mathrm{IDCG}},
\]
where $\mathrm{IDCG}$ is computed from the ideal descending order of $\mathbf{g}$.

\noindent\emph{MRR, Top-1, and Last-1 Accuracy.} Let $y^{\star}=\arg\max_{i} g_i$ be the ground-truth best candidate, and let $\mathrm{rank}_{\pi}(y^{\star})$ be its 1-based position in the predicted ranking. The mean reciprocal rank is
\[
\mathrm{MRR}=\frac{1}{\mathrm{rank}_{\pi}(y^{\star})}.
\]
Top-1 Accuracy measures whether the predicted best candidate matches the ground-truth best:
\[
\mathrm{Top1}
=
\mathbb{I}\!\left[\arg\max_{i}\hat{s}_{i}=\arg\max_{i}g_{i}\right].
\]
Similarly, the supplementary evaluation protocol also reports Last-1 Accuracy, defined as
\[
\mathrm{Last1}
=
\mathbb{I}\!\left[\arg\min_{i}\hat{s}_{i}=\arg\min_{i}g_{i}\right].
\]

\noindent\emph{Kendall's $\tau$ and Spearman's $\rho$.} Let $r^{\mathrm{pred}}_i$ and $r^{\mathrm{gt}}_i$ denote the predicted and ground-truth ranks of candidate $i$. We compute rank correlation using
\[
\tau
=
\frac{C-D}{\binom{K}{2}},
\]
where $C$ and $D$ are the numbers of concordant and discordant pairs, respectively, and
\[
\rho
=
\mathrm{corr}\!\left(\mathbf{r}^{\mathrm{pred}}, \mathbf{r}^{\mathrm{gt}}\right).
\]
In the standalone evaluation framework, these correlations are computed over the predicted and ground-truth rank indices. In the training objective, the same two statistics are computed directly from the predicted score vector and the ground-truth score vector; both serve to measure global rank consistency.

\noindent\emph{Pairwise Concordance.} During RL training, the reward module uses a dense pairwise concordance score over all valid pairs within the candidate set:
\[
\mathrm{Conc}(\hat{\mathbf{s}},\mathbf{g})
=
\frac{1}{|\mathcal{Q}|}
\sum_{1\le i<j\le K}
\mathbb{I}[g_i \neq g_j]
\cdot
\mathbb{I}\!\left[
\left(\hat{s}_{i}>\hat{s}_{j}\right)
\Leftrightarrow
\left(g_{i}>g_{j}\right)
\right],
\]
where $\mathcal{Q}=\{(i,j)\mid 1\le i<j\le K,\ g_i\neq g_j\}$. This quantity is the listwise analogue of PairAcc.

\noindent\emph{Global Ranking Reward.} The final reward is defined as a weighted combination of several ranking metrics:
\[
\mathcal{R}_{\mathrm{rank}}
=
\lambda_{\mathrm{ndcg}}\mathrm{NDCG}
+\lambda_{\tau}\widetilde{\tau}
+\lambda_{\rho}\widetilde{\rho}
+\lambda_{\mathrm{top1}}\mathrm{Top1}
+\lambda_{\mathrm{pair}}\mathrm{Conc}
+\lambda_{\mathrm{mrr}}\mathrm{MRR},
\]
where Kendall's $\tau$ and Spearman's $\rho$ are first normalized from $[-1,1]$ to $[0,1]$:
\[
\widetilde{\tau}=\frac{\tau+1}{2},\qquad
\widetilde{\rho}=\frac{\rho+1}{2}.
\]
In our experiments, we use
\[
(\lambda_{\mathrm{ndcg}},\lambda_{\tau},\lambda_{\rho},\lambda_{\mathrm{top1}},\lambda_{\mathrm{pair}},\lambda_{\mathrm{mrr}})
=(0.25,\,0.20,\,0.20,\,0.15,\,0.10,\,0.10).
\]
This reward encourages the policy to align not only with local pairwise preferences, but also with the global topology of the full candidate list.

\subsection{Training Details}
\noindent\emph{Model Initialization.} Our evaluator is built upon Qwen2.5-Omni-3B and replaces the original discrete language head with a one-layer continuous regression head. RL post-training is initialized from the pairwise preference-aligned checkpoint rather than from the untuned backbone, which substantially stabilizes early optimization and preserves the comparative structure learned during the cold-start stage. During optimization, the reference branch is kept frozen, while the trainable branch updates the regression head together with the selected transformer layers used for cross-modal reasoning.

\noindent\emph{Input Processing and Batch Construction.} All training and evaluation samples are standardized to 5-second clips, resized to $140\times 140$, and processed at 12 FPS. In the listwise RL setting, each training instance consists of one shared visual anchor together with $K=6$ candidate methods. To preserve the full listwise structure within each sample and avoid excessive memory growth from parallel multimodal forward passes, we use a per-device batch size of 1 during RL post-training. Validation is conducted independently from training with a small batch size for stable evaluation throughput.

\noindent\emph{Optimization Setup.} We optimize the model using AdamW with learning rate $1\times 10^{-6}$ and weight decay $5\times 10^{-2}$. The GRPO objective uses clipping radius $\epsilon=0.2$, KL regularization weight $\beta=10^{-3}$, rollout number $N=12$, and Gaussian exploration variance centered at $\sigma=2.5$. Training is performed for 100 epochs with gradient clipping at 1.0. Following the main experiments, validation is executed once per epoch, and the best checkpoint is selected according to validation pairwise accuracy.

\noindent\emph{Distributed Training.} RL post-training is carried out with mixed-precision distributed optimization using six GPUs under a memory-efficient sharded training strategy. This configuration is necessary because each listwise sample requires multiple multimodal forward passes, one for each candidate method, and the rollout-based policy optimization further increases memory pressure. The chosen setup allows the full listwise objective to be optimized without reducing the candidate set size or truncating rollout evaluation.

\noindent\emph{Reproducibility.} Unless otherwise stated, all experiments use a fixed random seed of 42. We additionally enable deterministic cuDNN behavior and disable benchmark auto-tuning to reduce nondeterminism across runs. Dataloaders are refreshed at the epoch level so that shuffled listwise groupings remain synchronized with the distributed training process. Taken together, these settings yield stable RL trajectories and reproducible validation performance across repeated runs.

\subsection{Computational Efficiency}
\noindent\emph{Inference Cost.} A practical evaluation metric must be cheap enough for large-scale deployment, particularly when used as a reward model during generative model training. Although our evaluator is built upon a 3B Omni-LLM, it computes a synchronization score from a \emph{single forward pass} over the multimodal context, reading the hidden state at the \texttt{[SCORE]} token without any autoregressive decoding. Table~\ref{tab:compute_cost} reports the parameter count, peak memory, latency, and FLOPs of representative evaluators measured on a single RTX~4090 under vanilla HuggingFace inference. Our metric runs at $0.23$\,s per pair with $10.0$G memory, which is comparable to lightweight CNN-based baselines and roughly $5\times$ faster and $6\times$ cheaper in FLOPs than the 8.3B LLM-as-Judge (VQ-Insight). This confirms that the substantial gains in human-preference alignment do not come at a prohibitive computational cost.

\begin{table}[t]
\centering
\caption{Computational cost of synchronization evaluators, measured on a single RTX~4090 with vanilla HuggingFace inference. Our 3B Omni-LLM requires only a single forward pass (no autoregressive decoding), keeping latency and FLOPs on par with lightweight CNN baselines and far below the 8.3B LLM-as-Judge.}
\label{tab:compute_cost}
\setlength{\tabcolsep}{8pt}
\renewcommand{\arraystretch}{1.1}
\begin{tabular}{lcccc}
\toprule
Method & Params & Mem & Lat.\,(s) & TFLOPs \\
\midrule
AV-Align        & \textless50M & \textless1G & 0.16 & --    \\
DeSync          & 237M         & 2.1G        & 0.11 & 5.72  \\
JavisScore      & 1.2B         & 6.4G        & 0.05 & 2.13  \\
VQ-Insight      & 8.3B         & 19.6G       & 1.08 & 62.49 \\
\textbf{Ours}   & 3B           & 10.0G       & 0.23 & 10.35 \\
\bottomrule
\end{tabular}
\end{table}

\subsection{RL Post-Training Pseudocode}
\noindent\emph{Algorithmic Summary.} Finally, we provide pseudocode for the RL stage of \method, summarizing the rollout construction, Gaussian score sampling, reward computation, and policy update procedure. This presentation makes explicit how continuous scalar scoring is reparameterized as a stochastic policy and optimized under the group-relative objective.

\begin{algorithm}[t]
\caption{RL Post-Training with Ranking-Based \method}
\label{alg:rl_grpo}
\small
\begin{algorithmic}[1]
\Require Policy model $\pi_{\theta}$ with scalar head, frozen reference model $\pi_{\mathrm{ref}}$, ranking batch $\mathcal{B}=\{(x_b,\mathbf{y}_b,\mathbf{g}_b)\}_{b=1}^{B}$, rollout count $N$, Gaussian standard deviation $\sigma$, clipping radius $\epsilon$, KL weight $\beta$
\For{each training iteration}
    \State Flatten all valid candidates in $\mathcal{B}$ and compute scalar scores
    \Statex \hspace{1.5em} $\mu_{\theta}^{(i)} \gets \pi_{\theta}(x^{(i)}, y^{(i)})$, \quad $\mu_{\mathrm{ref}}^{(i)} \gets \pi_{\mathrm{ref}}(x^{(i)}, y^{(i)})$
    \For{each candidate $i$}
        \For{$n=1$ to $N$}
            \State Sample rollout score $a^{(i,n)} \sim \mathcal{N}(\mu_{\theta}^{(i)}, \sigma^2)$
        \EndFor
    \EndFor
    \State Regroup rollout scores into sample-level candidate sets $\mathbf{a}_{b}^{(n)}=\{a_{b,1}^{(n)},\dots,a_{b,K_b}^{(n)}\}$
    \For{each sample $b$ and rollout $n$}
        \State Compute ranking reward $r_b^{(n)} \gets \mathcal{R}_{\mathrm{rank}}(\mathbf{a}_{b}^{(n)}, \mathbf{g}_b)$
    \EndFor
    \For{each sample $b$}
        \State $\bar{r}_b \gets \frac{1}{N}\sum_{n=1}^{N} r_b^{(n)}$, \quad $s_b \gets \mathrm{Std}(\{r_b^{(n)}\}_{n=1}^{N})$
        \For{$n=1$ to $N$}
            \State $A_b^{(n)} \gets \frac{r_b^{(n)} - \bar{r}_b}{s_b + 10^{-8}}$
        \EndFor
        \State Assign the same $A_b^{(n)}$ to all candidates belonging to sample $b$
    \EndFor
    \For{each candidate $i$ and rollout $n$}
        \State $\rho^{(i,n)} \gets \exp\!\left(\frac{(a^{(i,n)}-\mu_{\mathrm{ref}}^{(i)})^2-(a^{(i,n)}-\mu_{\theta}^{(i)})^2}{2\sigma^2}\right)$
    \EndFor
    \State $\mathcal{L}_{\mathrm{GRPO}} \gets - \mathbb{E}_{i,n}\left[\min\left(\rho^{(i,n)}A^{(i,n)}, \mathrm{clip}(\rho^{(i,n)},1-\epsilon,1+\epsilon)A^{(i,n)}\right)\right]$
    \State $\mathcal{L}_{\mathrm{KL}} \gets \mathbb{E}_{i}\left[\frac{\sigma^2+(\mu_{\theta}^{(i)}-\mu_{\mathrm{ref}}^{(i)})^2}{2\sigma^2}-\frac{1}{2}\right]$
    \State Update $\theta$ using $\mathcal{L} \gets \mathcal{L}_{\mathrm{GRPO}} + \beta \mathcal{L}_{\mathrm{KL}}$
\EndFor
\State \Return Updated synchronization evaluator $\pi_{\theta}$
\end{algorithmic}
\end{algorithm}

\end{document}